\documentclass{article}
\usepackage{multirow}
\usepackage{tabularx}
\usepackage{longtable}
     \PassOptionsToPackage{numbers, compress, sort}{natbib}

\usepackage{graphicx}
\usepackage{caption} 
\usepackage{subcaption} 
\usepackage{booktabs}
\usepackage{wrapfig}
\usepackage[colorinlistoftodos,prependcaption,textsize=tiny]{todonotes}

     \usepackage[final]{neurips_data_2021}

\usepackage[utf8]{inputenc} 
\usepackage[T1]{fontenc}    
\usepackage{hyperref}       
\usepackage{url}            
\usepackage{booktabs}
\usepackage{tabularx}       
\usepackage{amsfonts}       
\usepackage{nicefrac}       
\usepackage{microtype}      
\usepackage{xcolor}         

\usepackage{amsmath,amsfonts,bm}

\def\eqref#1{equation~\ref{#1}}

\def\1{\bm{1}}

\DeclareMathAlphabet{\mathsfit}{\encodingdefault}{\sfdefault}{m}{sl}
\SetMathAlphabet{\mathsfit}{bold}{\encodingdefault}{\sfdefault}{bx}{n}

\newcommand{\searchspace}{\mathcal{X}}
\newcommand{\dataset}{\mathcal{D}}
\newcommand{\conf}{\mathbf{x}}
\newcommand{\Conf}{\mathbf{X}}
\newcommand{\benchname}{HPO-B}

\usepackage{selectp}

\makeatletter
\newcommand{\subalign}[1]{%
  \vcenter{%
    \Let@ \restore@math@cr \default@tag
    \baselineskip\fontdimen10 \scriptfont\tw@
    \advance\baselineskip\fontdimen12 \scriptfont\tw@
    \lineskip\thr@@\fontdimen8 \scriptfont\thr@@
    \lineskiplimit\lineskip
    \ialign{\hfil$\m@th\scriptstyle##$&$\m@th\scriptstyle{}##$\hfil\crcr
      #1\crcr
    }%
  }%
}

\title{\benchname{}: A Large-Scale Reproducible Benchmark \\ for Black-Box HPO based on OpenML}

\author{%
  Sebastian Pineda Arango\thanks{Equal contribution} \\
  University of Freiburg\\
  \texttt{pineda@cs.uni-freiburg.de} \\
  	\And 
    Hadi S. Jomaa\footnotemark[1] \\
    University of Hildesheim \\
    {\tt hsjomaa@ismll.uni-hildesheim.de}
	
	\And 
    Martin Wistuba\thanks{Work done prior joining Amazon Research} \\
    Amazon Research \\
    {\tt marwistu@amazon.com}
    
    \And
	Josif Grabocka\\
  	University of Freiburg\\
	    {\tt grabocka@cs.uni-freiburg.de} \\

}

\begin{document}

\maketitle

\begin{abstract}
Hyperparameter optimization (HPO) is a core problem for the machine learning community and remains largely unsolved due to the significant computational resources required to evaluate hyperparameter configurations. 
As a result, a series of recent related works have focused on the direction of transfer learning for quickly fine-tuning hyperparameters on a dataset. Unfortunately, the community does not have a common large-scale benchmark for comparing HPO algorithms. Instead, the \textit{de facto} practice consists of empirical protocols on arbitrary small-scale meta-datasets that vary inconsistently across publications, making reproducibility a challenge. To resolve this major bottleneck and enable a fair and fast comparison of black-box HPO methods on a level playing field, we propose \textbf{HPO-B}, a new large-scale benchmark in the form of a collection of meta-datasets. Our benchmark is assembled and preprocessed from the OpenML repository and consists of 176 search spaces (algorithms) evaluated sparsely on 196 datasets with a total of 6.4 million hyperparameter evaluations. For ensuring reproducibility on our benchmark, we detail explicit experimental protocols, splits, and evaluation measures for comparing methods for both non-transfer, as well as, transfer learning HPO.
\end{abstract}

\section{Introduction}

Hyperparameter Optimization (HPO) is arguably the major open challenge for the machine learning community due to the expensive computational resources demanded to evaluate configurations.
As a result, HPO and its broader umbrella research area, AutoML, have drawn particular interest over the past decade~\cite{bergstra2011algorithms,springenberg2016bayesian,snoek2015scalable,hutter2011sequential}.
Black-box HPO is a specific sub-problem that focuses on the case where the function to be optimized (e.g. the generalization performance of an algorithm) is unknown, non-differentiable with respect to the hyperparameters, and intermediate evaluation proxies are not computable (opposed to gray-box HPO~\citep{li2017hyperband} which accesses intermediate performance measurements).

Although black-box HPO is a core problem, existing solutions based on parametric surrogate models for estimating the performance of a configuration overfit the limited number of evaluated configurations. As a result, the AutoML community has recently invested efforts in resolving the sample-inefficiency of parametric surrogates via meta- and transfer-learning~\cite{schilling2015hyperparameter,wistuba2016two, feurer2018scalable,volpp2019meta,jomaa2019hyp,wistuba2021few,salinas2020quantile}.

Unfortunately, despite the promising potential of transfer-learning in black-box HPO, the impact of such algorithms is hindered by their poor experimental reproducibility. Our personal prior research experience, as well as the feedback from the community, highlight that reproducing and generalizing the results of transfer-learning HPO methods is challenging. In essence, the problem arises when the results of a well-performing method in the experimental protocol of a publication either can not be replicated; or when the method underperforms in a slightly different empirical protocol. We believe that a way of resolving this negative \textit{impasse} is to propose a new public large-scale benchmark for comparing HPO methods, where the exact training/validation/test splits of the meta-datasets, the exact evaluation protocol, and the performance measures are well-specified. The strategy of adopting benchmarks is a trend in related areas, such as in computer vision~\citep{deng2009imagenet}, or NAS~\citep{pmlr-v97-ying19a,Dong2020NAS-Bench-201:}. 

In this perspective, we present \textbf{HPO-B}\footnote{The benchmark is publicly available at \url{https://github.com/releaunifreiburg/HPO-B}}, the largest public benchmark of meta-datasets for black-box HPO containing 6.4M hyperparameter evaluations across 176 search spaces (algorithms) and on 196 datasets in total. The collection is derived from the raw data of OpenML~\citep{OpenML2013}, but underwent an extensive process of cleaning, preprocessing and organization (Section~\ref{sec:compandprep}). Additionally, we offer off-the-shelf ready variants of the benchmark that are adapted for both non-transfer, as well as transfer HPO experiments, together with the respective evaluation protocols (Section~\ref{sec:expprotocol}). This large, diverse, yet plug-and-play benchmark can significantly boost future research in black-box HPO.

\section{Terminology}
To help the reader navigate through our paper, we present the compact thesaurus of Table~\ref{tab:terminology} for defining the vernacular of the HPO community.

\begin{table}[ht]
\centering
\begin{tabular}{ll}
\toprule
\textbf{Term} & \textbf{Definition} \\
\midrule
Configuration & Specific settings/values of hyperparameters \\
Search space & The domain of a configuration: scale and range of each hyperparameter's values \\
Response & The performance of an algorithm given a configuration and dataset\\
Surrogate & A (typically parametric) function that approximates the response\\
Seed & Set of initial configurations used to fit the initial surrogate model\\
Black-box & The response is an unknown and non-differentiable function of a configuration\\
Task & An HPO problem given a search space and a dataset \\
Evaluation & The measured response of a configuration on a dataset \\
Trial & An evaluation on a task during the HPO procedure \\
Meta-dataset & Collection of \textit{recorded} evaluations from different tasks on a search space \\
Meta-instance & An evaluation in the meta-dataset for one of the tasks \\
Meta-feature & Descriptive attributes of a dataset \\
Source tasks & In a meta- or transfer-learning setup refers to the \textit{known} tasks we \textit{train from} \\
Target tasks & In a meta- or transfer-learning setup refers to the \textit{new} tasks we \textit{test on} \\
Benchmark & \textbf{New definition:} Collection of meta-datasets from different search spaces \\
\bottomrule
\end{tabular}\\
\caption{\label{tab:terminology}A thesaurus of the common HPO terminology used throughout this paper
}
\end{table}

\section{Related Work}
\label{sec:relatedwork}

\textbf{Non-transfer black-box HPO}: The mainstream paradigm in HPO relies on surrogates to estimate the performance of hyperparameter configurations. For example, \cite{bergstra2011algorithms} were the first to propose Gaussian Processes (GP) as surrogates. The same authors also propose a Tree Parzen Estimator (TPE) for computing the non-parametric densities of the hyperparameters given the observed performances. Both approaches achieve a considerable lift over random~\cite{bergstra2012random}  and manual search. To address the cubic run-time complexity of GPs concerning the number of evaluated configurations, DNGO~\cite{snoek2015scalable} trains neural networks for generating adaptive basis functions of hyperparameters, in combination with a Bayesian linear regressor that models uncertainty. Alternatively, SMAC~\cite{hutter2011sequential} represents the surrogate as a random forest, and BOHAMIANN~\cite{springenberg2016bayesian} employs Bayesian Neural Networks instead of plain neural networks to estimate the uncertainty of a configuration's performance. For an extensive study on non-transfer Bayesian Optimization techniques for HPO, we refer the readers to \cite{snoek2012practical,Cowen2020empirical} that study the impact of the underlying assumptions associated with black-box HPO algorithms.

\textbf{Transfer black-box HPO}: To expedite HPO, it is important to leverage information from existing evaluations of configurations from prior tasks. A common approach is to capture the similarity between datasets using meta-features (i.e. descriptive dataset characteristics). Meta-features have been used as a warm-start initialization technique~\cite{feurer2015initializing,jomaa2021dataset2vec}, or as part of the surrogate directly~\cite{bardenet2013collaborative}. Transfer learning is also explored through the weighted combination of surrogates, such as in TST-R~\cite{wistuba2016two}, RGPE~\cite{feurer2018scalable}, and TAF-R~\cite{wistuba2018scalable}. Another direction is learning a shared surrogate across tasks. ABLR optimizes a shared hyperparameter embedding with separate Bayesian linear regressors per task~\cite{perrone2018scalable}, while GCP~\cite{salinas2020quantile} maps the hyperparameter response to a shared distribution with a Gaussian Copula process. Furthermore, FSBO~\cite{wistuba2021few}
meta-learns a deep-kernel Gaussian Process surrogate, whereas DMFBS incorporates the dataset context through end-to-end meta-feature networks~\cite{jomaa2021dataset2vec}.


\textbf{Meta-datasets}: The work by Wistuba et al. \cite{wistuba2015learning} popularised the usage of meta-dataset benchmarks with pre-computed evaluations for the hyperparameters of SVM (288 configurations) and Adaboost (108 configurations) on 50 datasets; a benchmark that inspired multiple follow-up works~\cite{feurer2018scalable,volpp2020metalearning}.
Existing attempts to provide HPO benchmarks deal only with the non-transfer black-box HPO setup~\cite{eggensperger2013towards}, or the gray-box HPO setup~\cite{Katharina2021hpobench}.
As they contain results for one or very few datasets per search space, they cannot be used for the evaluation of transfer black-box HPO methods.
Nevertheless, there is a trend in using evaluations of search spaces from the OpenML repository~\cite{OpenML2020}, which contains evaluations reported by an open community, as well as large-scale experiments contributed by specific research labs~\cite{Kuhn2018_Automatic, Binder2020_Collecting}. However, the choice of OpenML search spaces in publications is ad-hoc: one related work uses SVM and XGBoost~\cite{perrone2018scalable}, a second uses GLMNet and SVM~\cite{wistuba2021few}, while a third paper uses XGBoost, Random Forest and SVM~\cite{perrone2019learning}. We assess that the community \textit{(i)} inconsistently cherry-picks (assuming \textit{bona fides}) search spaces, with \textit{(ii)} arbitrary train/validation/test splits of the tasks within the meta-dataset, and \textit{(iii)} inconsistent preprocessing of hyperparameters and responses. In our experiments, we observed that existing methods do not generalize well on new meta-datasets.

\textbf{Our Novelty:} As a remedy, we propose a novel benchmark derived from OpenML~\cite{OpenML2020}, that resolves the existing reproducibility issues of existing non-transfer and transfer black-box HPO methods, by ensuring a fairly-reproducible empirical protocol. The contributions of our benchmark are multi-fold. First of all, we remove the confounding factors induced by different meta-dataset preprocessing pipelines (e.g. hyperparameter scaling and transformations, missing value imputations, one-hot encodings, etc.). Secondly, we provide a specified collection of search spaces, with specified datasets and evaluations. Furthermore, for transfer learning HPO methods, we also provide pre-defined training/validation/testing splits of tasks. For experiments on the test tasks, we additionally provide 5 seeds (i.e. 5 sets of initial hyperparameters to fit the initial surrogate) with 5 hyperparameter configurations, each. We also highlight recommended empirical measures for comparing HPO methods and assessing their statistical significance in Section~\ref{sec:expprotocol}. In that manner, the results of different papers that use our benchmark can be compared directly without fearing the confounding factors. Table~\ref{tab:priormetadatasets} presents a summary of the descriptive statistics of meta-datasets from prior literature. To the best of our awareness, the proposed benchmark is also richer (in the number of search spaces and their dimensionality) and larger (in the number of evaluations) than all the prior protocols.

\begin{table}[ht]
\centering
\begin{tabular}{cccccr}
\toprule
\textbf{Paper} & \textbf{Venue/Year} & \textbf{\# Search Spaces} & \textbf{\# Datasets} & \textbf{\# HPs} & \textbf{\# Evals.} \\
\midrule
\cite{bardenet2013collaborative} & ICML '13 & 1 & 29 & 2 & 3K \\
\cite{wistuba2015learning} & DSAA '15 & 2 & 50 & 2, 4 & 20K \\
\cite{feurer2015initializing} & AAAI '15 & 3 & 57 & 4, 5 & 93K \\
\cite{wistuba2015hyperparameter} & ECML-PKDD '15 & 17 & 59 & 1-7 & 1.3M \\
\cite{perrone2018scalable} & NeurIPS '18 & 2 & 30 & 4, 10 & 655K \\
\cite{salinas2020quantile} & ICML '20 & 4 & 26 & 6, 9 & 343K \\
\cite{jomaa2021dataset2vec} & DMKD '21 & 1 & 120 & 7 & 414K \\
\cite{wistuba2021few} & ICLR '21 & 3 & 80 & 2, 4 & 864K \\
\midrule
Our \benchname{}-v1 & - & 176 & 196 & 1-53 & 6.39M \\
Our \benchname{}-v2/-v3 & - & 16 & 101 & 2-18 & 6.34M \\
\bottomrule
\end{tabular}\\
\caption{\label{tab:priormetadatasets}Summary statistics for various meta-datasets considered in prior works.
}

\end{table}

\section{A Brief Explanation of Bayesian Optimization Concepts}

As we often refer to HPO methods, in this section we present a brief coverage of Bayesian Optimization as the most popular HPO method for black-box optimization.
HPO aims at minimizing the function $f:\searchspace\rightarrow\mathbb{R}$ which maps each hyperparameter configuration $\conf \in \searchspace$ to the validation loss obtained when training the machine learning model using $\conf$.
Bayesian Optimization keeps track of all evaluated hyperparameter configurations in a history $\dataset=\{(\conf_i,y_i)\}_i$, where $y_i\sim\mathcal{N}(f(\conf_i), \sigma_n^2)$ is the (noisy) response which can be heteroscedastic~\cite{Griffiths2019achieving} in real-world problems~\cite{Cowen2020empirical}.
A probabilistic model, the so-called surrogate model, is used to approximate the behavior of the response function. Gaussian Processes are a common choice for the surrogate model~\cite{Rasmussen2006, snoek2012practical}. Bayesian Optimization is an iterative process that alternates between updating the surrogate model as described above and selecting the next hyperparameter configuration.
The latter is done by finding the configuration which maximizes an acquisition function, which scores each feasible hyperparameter configuration using the surrogate model by finding a trade-off between exploration and exploitation.
Arguably, the most popular acquisition function is the Expected Improvement~\cite{Jones1998_Efficient}.
The efficiency of Bayesian Optimization depends on the surrogate model's ability to approximate the response function.
However, this is a challenging task since every optimization starts with no or little knowledge about the response function. 
To overcome this cold-start problem, transfer methods have been proposed, which leverage information from other tasks of the same search space.

\section{Benchmark Description}
\label{sec:compandprep}

The benchmark \benchname{} is a collection of meta-datasets collected from OpenML~\citep{OpenML2020} with a diverse set of search spaces. We present three different versions of the benchmark, as follows:

\begin{itemize}
    \item \textbf{\benchname{}-v1:} The raw benchmark of all 176 meta-datasets;
    \item \textbf{\benchname{}-v2:} Subset of 16 meta-datasets with the most frequent search spaces;
    \item \textbf{\benchname{}-v3:} Split of \benchname{}-v2 into training, validation and testing.
\end{itemize}

When assembling the benchmark \benchname{}-v1 we noticed that most of the evaluations are reported for a handful of popular search spaces, in particular, we noticed that 9\% of the top meta-datasets include 99.3\% of the evaluations. As a result, we created a second version \benchname{}-v2 that includes only the frequent meta-datasets that have at least 10 datasets with at least 100 evaluations per dataset (Section~\ref{sec:metadatasummary}). Furthermore, as we clarified in Section~\ref{sec:relatedwork} a major reproducibility issue of the related work on transfer HPO is the lack of clear training, validation, and test splits for the meta-datasets. To resolve this issue, we additionally created \benchname{}-v3 as a derivation of \benchname{}-v2 with pre-defined splits of the training, validation, and testing tasks for every meta-dataset, in addition to providing initial configurations (seeds) for the test tasks. The three versions were designed to fulfill concrete purposes with regards to different types of HPO methods. For non-transfer black-box HPO methods, we recommend using \benchname{}-v2 which offers a large pool of HPO tasks. Naturally, for transfer HPO tasks we recommend using \benchname{}-v3 where meta-datasets are split into training, validation, and testing. We still are releasing the large \benchname{}-v1 benchmark to anticipate next-generation methods for heterogeneous transfer learning techniques that meta-learn surrogates across different search spaces, where all 176 meta-datasets might be useful despite most of them having few evaluations. 

Concretely, \benchname{}-v3 contains the set of filtered search spaces of \benchname{}-v2, which are specially split into \emph{four} sets: meta-train, meta-validation. meta-test and an augmented version of the meta-train dataset. Every split contains different datasets from the same search space. We distributed the datasets per search space as 80\% of the datasets to meta-train, 10\% to meta-validation, and 10\% to meta-test, respectively. A special, augmented version of the meta-train is created by adding all other search space evaluations from \benchname{}-v1 that are not part of \benchname{}-v3. On the other hand, in \benchname{}-v3 we also provide seeds for initializing the HPO. They are presented as five different sets of five initial configurations to be used by a particular HPO method. By providing five different seeds we decrease the random effect of the specific initial configurations. To ease the comparison among HPO methods, we suggest using the recommended initial configurations for testing. Although, we admit that some algorithms proposing novel warm-starting strategies might need to bypass the recommended initial configurations.

\subsection{Benchmark summary}
\label{sec:metadatasummary}

The created benchmark contains 6,394,555 total evaluations across 176 search spaces that are sparsely evaluated on 196 datasets.
By accounting for the search spaces that comply with our filtering criteria (at least 10 datasets with 100 evaluations), we obtain \benchname{}-v2 with 16 different search spaces and 6,347,916 evaluations on 101 datasets.
Notice that the benchmark does not include evaluations for all datasets in every search space.
The number of dimensions, datasets, and evaluations per search space is listed in Table~\ref{table:summary}.
An additional description of the rest of all the search spaces in \benchname{}-v1 is presented in the Appendix.
In addition, Table~\ref{table:summary} shows the description of the meta-dataset splits according to the \benchname{}-v3.

\begin{table}[ht]
\centering
\begin{tabular}{lcrrcrcrc}
\toprule
\multicolumn{1}{c}{\multirow{2}{*}{\textbf{Search Space}}} & \multirow{2}{*}{\textbf{ID}} & \multirow{2}{*}{\textbf{\#HPs}} & \multicolumn{2}{c}{\textbf{Meta-Train}} & \multicolumn{2}{c}{\textbf{Meta-Validation}} & \multicolumn{2}{c}{\textbf{Meta-Test}} \\ \cline{4-9} 
\multicolumn{1}{c}{}                                &            &                     & \textbf{\#Evals.}    &  \textbf{\#DS}   & \textbf{\#Evals.}      &  \textbf{\#DS}      &  \textbf{\#Evals.}   & \textbf{\#DS}   \\  \midrule
rpart.preproc (16)   &  4796                                               & 3                               & 10694            & 36           & 1198               & 4               & 1200            & 4            \\ 
svm (6)  &   5527                                                  & 8                               & 385115           & 51           & 196213             & 6               & 354316          & 6            \\ 
rpart (29)       &   5636                                                  & 6                               & 503439           & 54           & 184204             & 7               & 339301          & 6            \\ 
rpart (31)       &   5859                                                  & 6                               & 58809            & 56           & 17248              & 7               & 21060           & 6            \\ 
glmnet (4)       &   5860                                                  & 2                               & 3100             & 27           & 598                & 3               & 857             & 3            \\ 
svm (7)          &   5891                                                  & 8                               & 44091            & 51           & 13008              & 6               & 17293           & 6            \\ 
xgboost (4)      &   5906                                                  & 16                              & 2289             & 24           & 584                & 3               & 513             & 2            \\ 
ranger (9)       &   5965                                                  & 10                              & 414678           & 60           & 73006              & 7               & 83597           & 7            \\ 
ranger (5)       &   5970                                                  & 2                               & 68300            & 55           & 18511              & 7               & 19023           & 6            \\ 
xgboost (6)      &   5971                                                  & 16                              & 44401            & 52           & 11492              & 6               & 19637           & 6            \\ 
glmnet (11)      &   6766                                                  & 2                               & 599056           & 51           & 210298             & 6               & 310114          & 6            \\ 
xgboost (9)      &   6767                                                  & 18                              & 491497           & 52           & 211498             & 7               & 299709          & 6            \\ 
ranger (13)       &   6794                                                  & 10                              & 591831           & 52           & 230100             & 6               & 406145          & 6            \\ 
ranger (15)      &   7607                                                  & 9                               & 18686            & 58           & 4203               & 7               & 5028            & 7            \\ 
ranger (16)      &   7609                                                  & 9                               & 41631            & 59           & 8215               & 7               & 9689            & 7            \\ 
ranger (7)       &   5889                                                  & 6                               & 1433             & 20           & 410                & 2               & 598             & 2            \\ \bottomrule
\end{tabular}\\
\caption{Description of the search spaces in \benchname{}-v3; "\#HPs" stands for the number of hyperparameters, "\#Evals." for the number of evaluations in a search space, while "\#DS" for the number of datasets across which the evaluations are collected. The search spaces are named with the respective OpenML version number (in parenthesis), and their original names are preceded by \textit{mlr.classif.}}
\label{table:summary}
\end{table}

\subsection{Preprocessing}

The OpenML-Python API~\cite{feurer-arxiv19a} was used to download the experiment data from OpenML~\cite{OpenML2020}.
We have collected all evaluations (referred to as runs in OpenML) tagged with \texttt{Verified\_Supervised\_Classification} available until April 15, 2021.

While the hyperparameter configuration was directly available for many evaluations, some of them had to be parsed from WEKA arguments (e.g. \texttt{weka.filters.unsupervised. attribute.RandomProjection -P 16.0 -R 42 -D Sparse1)}.
A small percentage (<0.001\%) of these were too complex in structure to be automatically parsed, so they were discarded.
Duplicate responses for the same hyperparameter configuration have been resolved by keeping only one random response.
Finally, all tasks with less than five observations were also discarded.

All categorical hyperparameters were one-hot encoded, taking into account all categories that occur in the different datasets for a search space.
Missing values have been replaced with zeros and the corresponding missing indicator (a new feature) has been set to one.
Hyperparameters that had the same value for all configurations in a search space were dropped.
We manually decided which hyperparameters required log-scaling by inspecting the distributions of each hyperparameter in each space (considerable manual effort).
Finally, the hyperparameter ranges were scaled to $[0,1]$. Further details on the pre-processing are explained in Appendix \ref{section:further_preprocessing}.

\subsection{Benchmark JSON schema}
\label{section:meta_dataset_format}
 The benchmark is offered as easily accessible JSON files. The first-level key of each JSON schema corresponds to the search space ID, whereas the second-level key specifies the dataset ID.
 By accessing the JSON schema with the search space $s$ and the dataset $t$, we obtain the meta-dataset $\dataset^{(s,t)}=\{(\conf^{(s,t)}_i,y^{(s,t)}_i)\}_i$, $\conf^{(s,t)}_i\in\searchspace^{(s)}$.
 The meta-dataset exhibits the following structure, where $N$ denotes the number of evaluations available for the specific task:
 \begin{center}
   \texttt{ \{search\_space\_ID: \{dataset\_ID:\{$\Conf$:[[$\conf_1$],$\dots$,[$\conf_N $]], $y$:[[$y_1$],$\dots$,[$y_N$]]\}\}\} }
\end{center}


The initialization seeds are similarly provided as a JSON schema, where the third-level subschema has 5 keys whose values are the indices of the samples to use as initial configurations.

\subsection{An additional continuous variant of \benchname{}}
\label{section:benchmark_continuous}

OpenML~\citep{OpenML2020} offers only discrete evaluations of hyperparameter configurations. Continuous HPO search methods are not applicable out-of-the-box on the discrete meta-datasets of \benchname{}, because evaluations are not present for every possible configuration in a continuous space. To overcome this limitation, we release an additional continuous version of \benchname{} based on task-specific surrogates. For every task, we fit an XGBoost~\citep{Chen:2016:XST:2939672.2939785} regression model with a maximum depth of 6 ańd two cross-validated hyper-hyperparameters, concretely the learning rate and the number of rounds. We train the surrogates to approximate the observed response values of the evaluated configurations on each task. As a result, for any arbitrary configuration in the continuous space, the approximate evaluation of a configuration's response is computed through the estimation of the respective task's surrogate. Furthermore, a download link to the trained surrogate models is also provided in the repository  \footnote{\url{https://github.com/releaunifreiburg/HPO-B}}. 


\section{Recommended Experimental Protocol}
\label{sec:expprotocol}

One of the primary purposes of \benchname{} is to standardize and facilitate the comparison between HPO techniques on \textit{a level playing field}. In this section, we provide two specific recommendations: which benchmark to use for a type of algorithm and what metrics to use for comparing results.


\paragraph{Evaluation Metrics}
We define the average normalized regret at trial $e$ (a.k.a. average distance to the minimum) as $\min_{x\in\mathcal{X}^{(s,t)}_e} \left(f^{(s,t)}(x)-y^*_\text{min}\right)/\left(y^*_\text{max} - y^*_\text{min}\right)$ with $\mathcal{X}^{(s,t)}_e$ as the set of hyperparameters that have been selected by a
HPO method up to trial $e$, with $y^*_\text{min}$ and $y^*_\text{max}$ as the best and worst responses, respectively.
The average rank represents the mean across tasks of the ranks of competing methods computed using the test accuracies of the best configuration until the $e$-th trial.
Results across different search spaces are computed by a simple mean over the search-space-specific results.


\paragraph{Non-Transfer Black-Box HPO} Methods should be compared on
 all the tasks in \benchname{}-v2 and for each of the five initial configurations. The authors of future papers should report the normalized regret and the mean ranks for all trials from 1 to 100 (excluding the seeds). We recommend that the authors show both aggregated and per search-space (possibly moved to the appendix) mean regret and mean rank curves for trials ranging from 1 to 100. In other words, as many runs as the number of tasks for given space times the number of initialization seeds. To assess the statistical significance of methods, we recommend that critical difference diagrams \cite{Demsar2006statistical} be computed for the ranks of all runs @25, @50, and @100 trials.
 
\paragraph{Transfer Black-Box HPO} Methods should be compared on the meta-data splits contained in \benchname{}-v3. All competing methods should use exactly the evaluations of the provided meta-train datasets for meta- and transfer-learning their method, and tune the hyper-hyperparameters on the evaluations of the provided meta-validation datasets. In the end, the competing methods should be tested on the provided evaluations of the meta-test tasks. As our benchmark does not have pre-computed responses for all possible configurations in a space, the authors either \textit{(i)} need to adapt their HPO acquisitions and suggest the next configuration only from the set of the pre-computed configurations for each specific meta-test task, or \textit{(ii)} use the continuous variant of HPO-B. Additionally, we recommend that the authors present (see details in the paragraph above) regret and rank plots, besides the critical difference diagrams @25, @50, and @100 trials. If a future transfer HPO method proposes a novel strategy for initializing configurations, for the sake of reproducibility we still recommend showing additional results with our initial configurations.


\begin{figure}[t]
\centering
\begin{subfigure}[b]{0.58\textwidth}
    \includegraphics[width=1.0\textwidth]{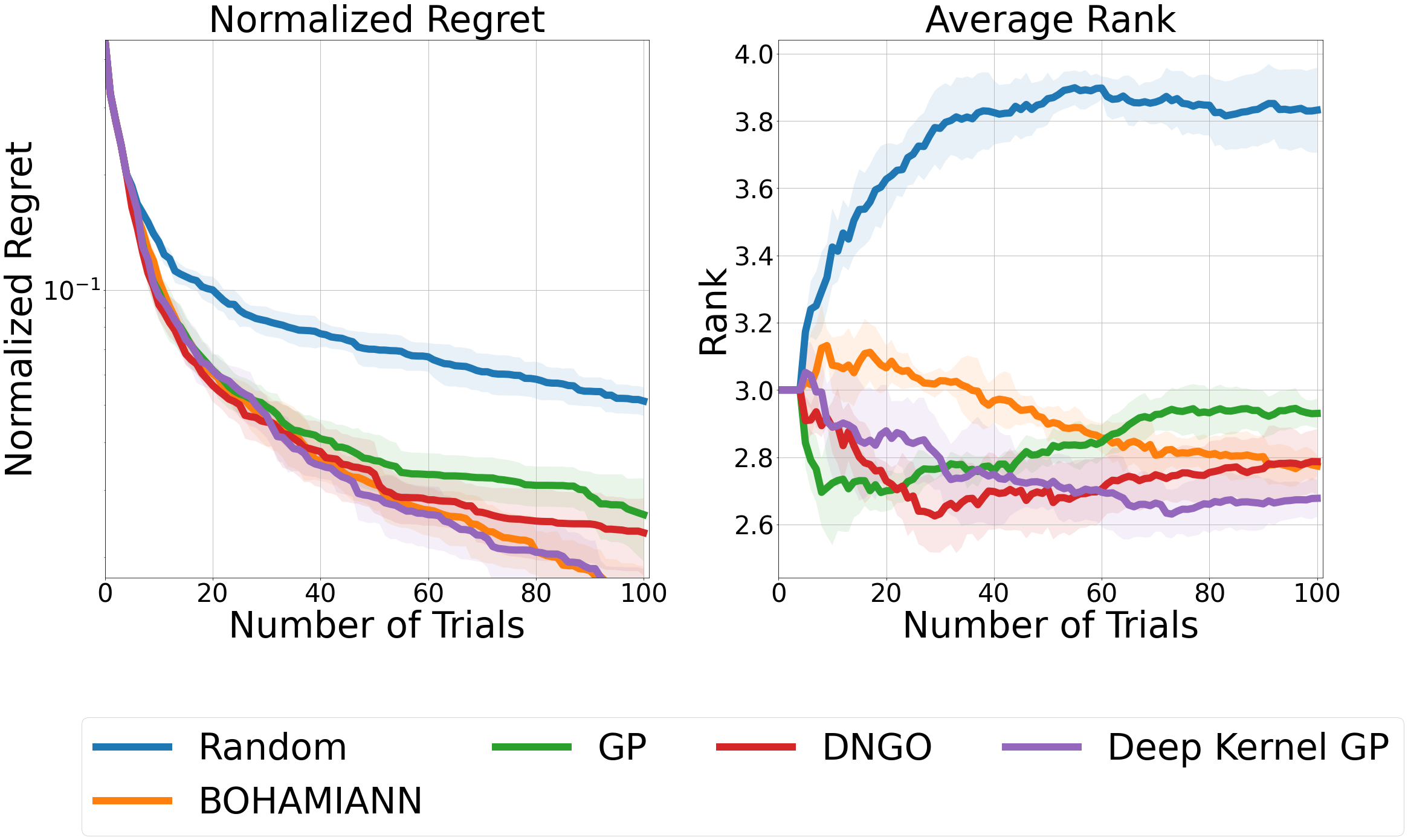}
\end{subfigure} 
\hfill
\begin{subfigure}[b]{0.41\textwidth}
\includegraphics[width=1.0\textwidth]{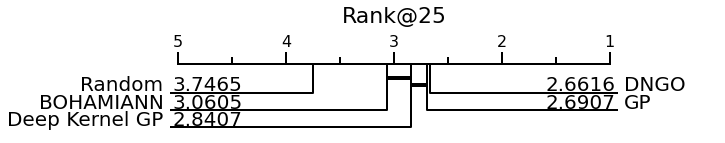}
\medskip
\includegraphics[width=1.0\textwidth]{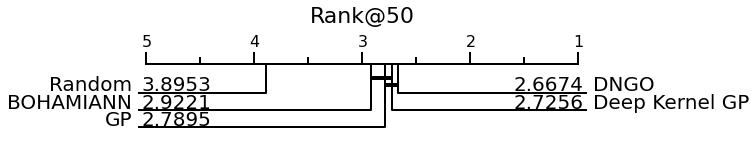}
\medskip
\includegraphics[width=1.0\textwidth]{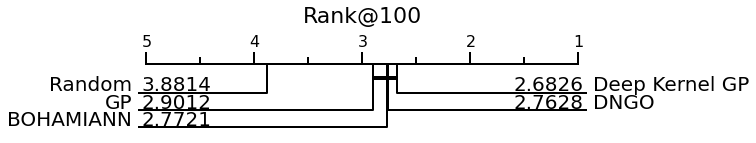}
\end{subfigure}
\caption{\textbf{Aggregated} comparisons of normalized regret and mean ranks across all search spaces for the \textbf{non-transfer} HPO methods on \benchname{}-v2}
    \label{fig:averagenontransfer}
\end{figure}

\section{Experimental Results}
\label{sec:results}

The benchmark is intended to serve as a new standard for evaluating non-transfer and transfer black-box HPO methods.
In the following, we will compare different methods according to our recommended protocol described in Section~\ref{sec:expprotocol}.
This is intended to demonstrate the usefulness of our benchmark, while at the same time serving as an example for the aforementioned recommendations on comparing baselines and presenting results.

\begin{figure}[htb]
\centering
\begin{subfigure}[b]{0.58\textwidth}
    \includegraphics[width=1.0\textwidth]{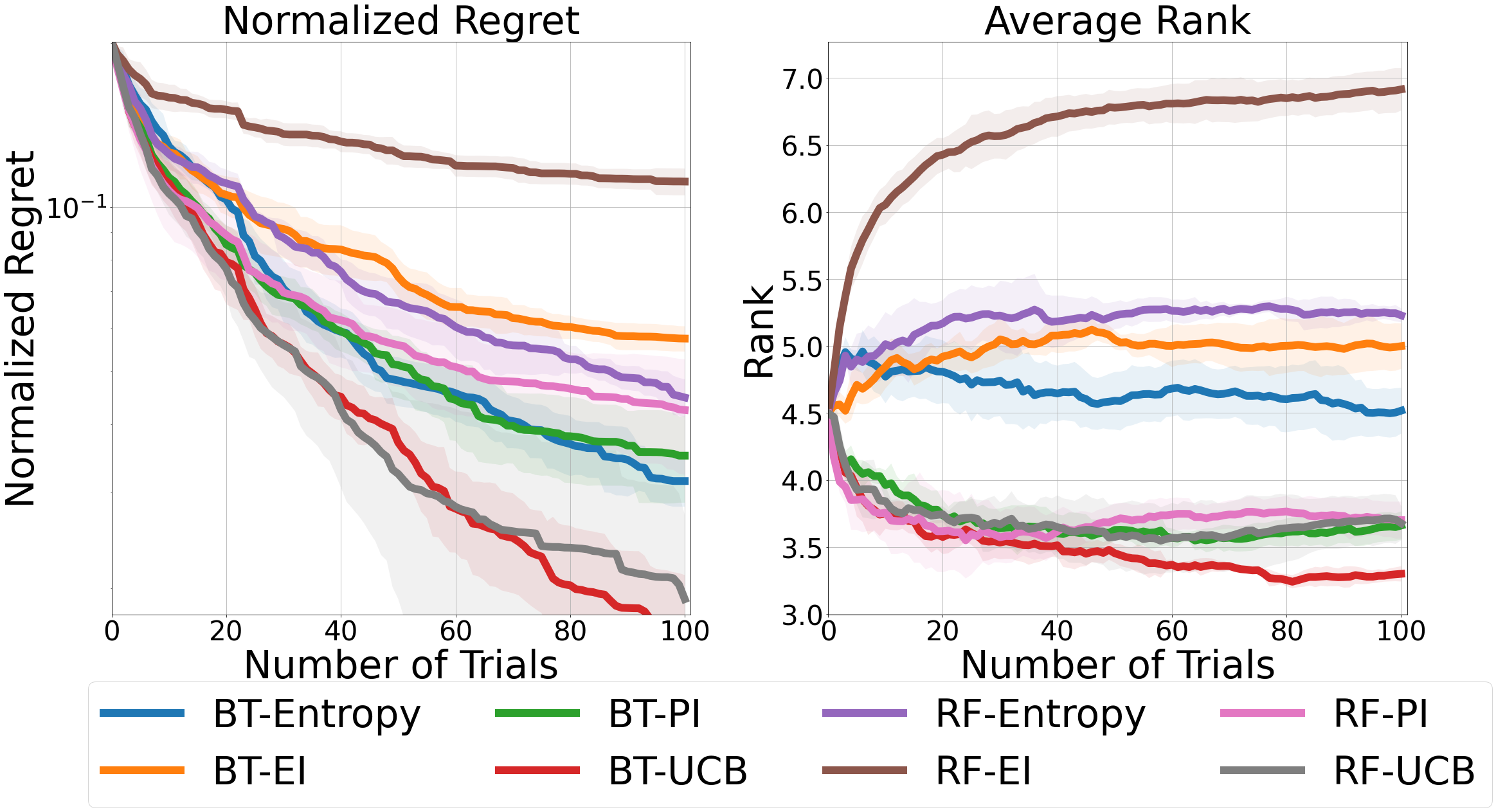}
\end{subfigure} 
\begin{subfigure}[b]{0.41\textwidth}
\includegraphics[width=1.0\textwidth]{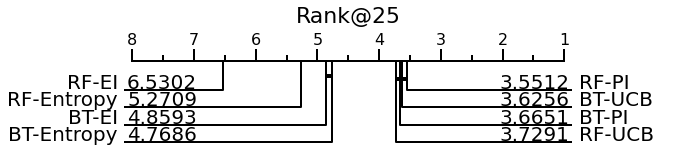}
\medskip
\includegraphics[width=1.0\textwidth]{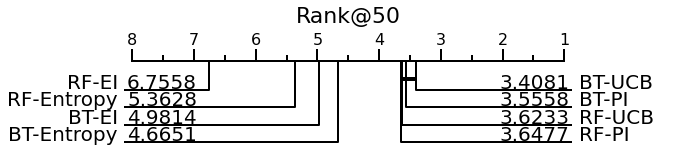}
\medskip
\includegraphics[width=1.0\textwidth]{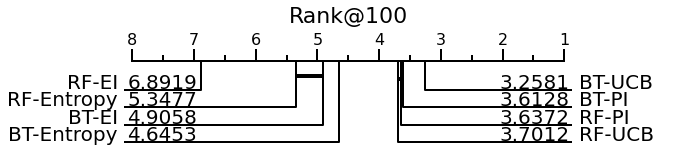}
\end{subfigure}
\caption{\textbf{Aggregated} comparisons of different surrogates and acquisition functions for \textbf{transfer} HPO methods on \benchname{}-v2; BT stands for Boosted Trees, RF for Random Forests, EI for Expected Improvement, and UCB for Upper Confidence Bound.}
    \label{fig:averageablation}
\end{figure}

\subsection{Non-transfer Black-Box HPO}

First, we compare Random Search, DNGO, BOHAMIANN, Gaussian Process (GP) with Mat\'{e}rn 3/2 kernel, and Deep Gaussian Process (FSBO~\cite{wistuba2021few} without pre-training) on \benchname{}-v2 in the non-transfer scenario.
As recommended by us earlier, in Figure~\ref{fig:regretnontransfer} we report aggregated results for normalized regret, average rank, and critical difference plots.
In addition, we report in Figure~\ref{fig:averagenontransfer} the aggregated normalized regret per search space.
The values in the figures for the number of trials equal to 0 correspond to the result after the five initialization steps. According to Figure~\ref{fig:averagenontransfer}, BOHAMIANN and Deep GP achieve comparable aggregated normalized regret across all search spaces, which suggests that both methods are equally well-suited for the tasks.
The average rank and the critical difference plot paint a different picture, in which Deep GP and DNGO achieve better results.
This discrepancy arises because each metric measures different performance aspects on different tasks, so it's important to report both.
As can be seen in Figure~\ref{fig:ranknontransfer}, Deep GP achieves better results than the GP in most of the tasks, which leads to a better average ranking.
However, as we can see in Figure~\ref{fig:regretnontransfer}, the regrets are observed at heterogeneous scales that can skew the overall averages. In some cases where BOHAMIANN outperforms Deep GP (e.g. search spaces 5527, 5859, and 5636), the difference in normalized regret is evident, due to the nature of the search space, whereas in cases where it is the other way around, however, the difference is only slightly less evident (e.g. search spaces 4796, 5906, and 7609). An important aspect of HPO is the choice of the surrogate function and acquisition.  Figure~\ref{fig:averageablation} presents an ablation of typical combinations and shows the accuracy of the Boosted Tree as a surrogate.

\begin{figure}[htb!]
\centering
\begin{subfigure}[b]{0.58\textwidth}
      \centering
      \includegraphics[width=1.0\textwidth]{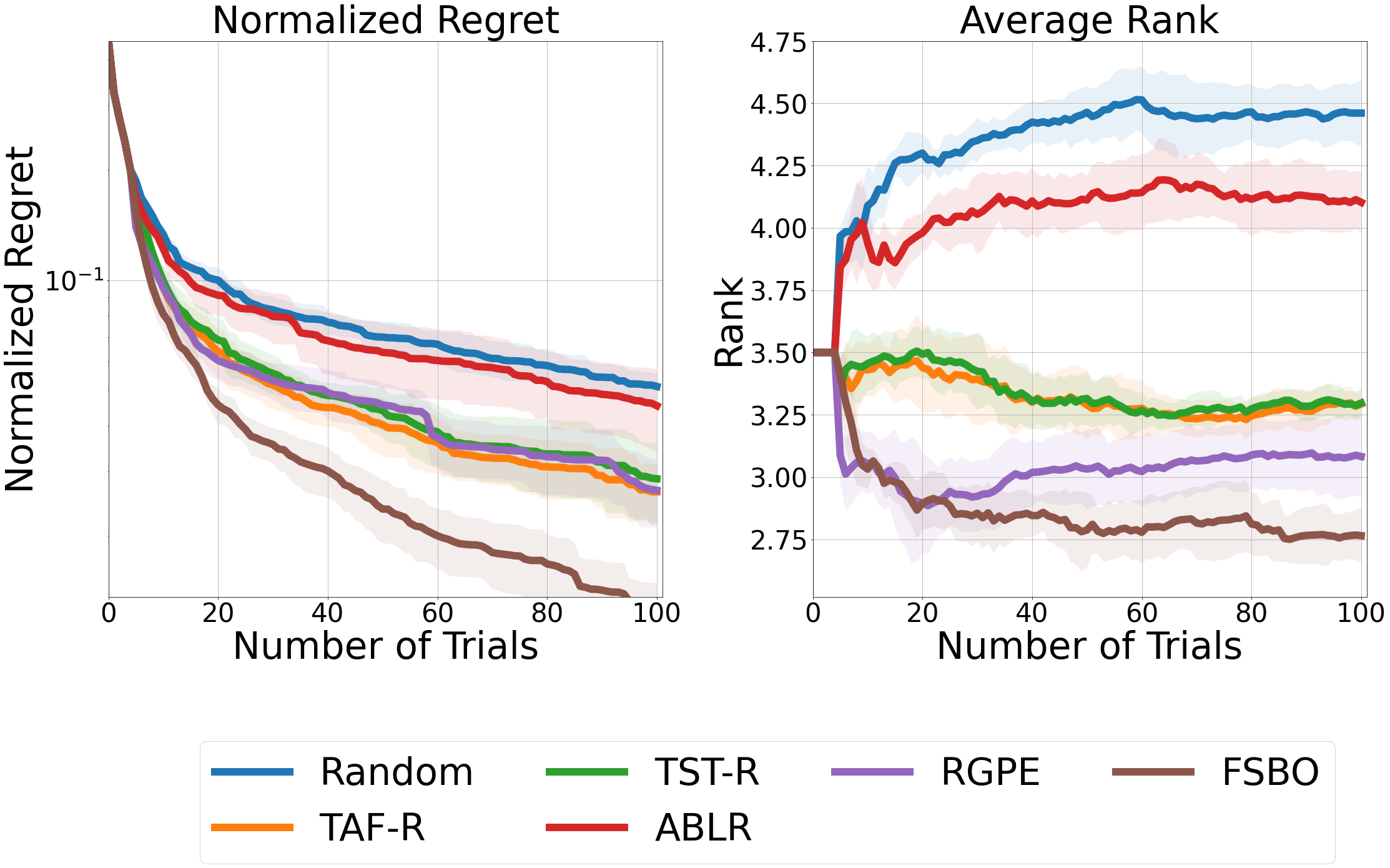}
\end{subfigure} 
\begin{subfigure}[b]{0.41\textwidth}
\includegraphics[width=1.0\textwidth]{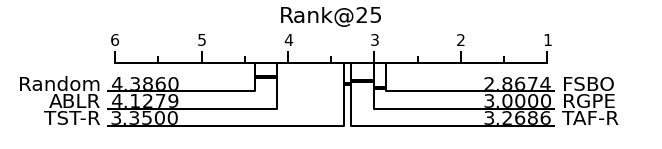}
\medskip
\includegraphics[width=1.0\textwidth]{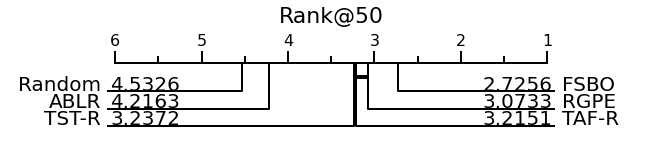}
\medskip
\includegraphics[width=1.0\textwidth]{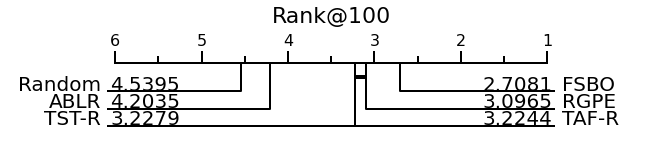}
\end{subfigure}
 \caption{\textbf{Aggregated} comparisons of normalized regret and mean ranks across all search spaces for the \textbf{transfer learning} HPO methods on \benchname{}-v3}
    \label{fig:averagetransfer}
\end{figure}

\subsection{Transfer Black-Box HPO}

Finally, we compare RGPE~\cite{feurer2018scalable}, ABLR~\cite{perrone2018scalable}, TST-R \cite{wistuba2016two}, TAF-R~\cite{wistuba2018scalable}, and FSBO \cite{wistuba2021few} on \benchname{}-v3 in the transfer scenario.
All hyper-hyperparameters were optimized on the meta-validation datasets and we report results aggregated across all test search spaces in terms of normalized regret and average rank in Figure~\ref{fig:averagetransfer}.
The results per search space for normalized regret and average rank are given in Figure~\ref{fig:regrettransfer} and Figure~\ref{fig:ranktransfer}, respectively. FSBO shows improvements over all the compared methods for the normalized regret metric and average rank metric. On the other hand, RGPE is seemingly performing similar to TST-R and TAF-R for the average regret, but performs significantly better for the average rank metric.
The explanation is the same as for our last experiment and can mainly be traced back to the strong performance of RGPE in search spaces 5971 and 5906. Such behaviors strengthen our recommendations of Section~\ref{sec:expprotocol} for showing results in terms of both the ranks and the normalized regrets, as well as the ranks' statistical significance.

\subsection{Comparing Non-Transfer vs. Transfer Black-Box HPO}

\begin{figure}[htb!]
\centering
\begin{subfigure}[b]{0.58\textwidth}
      \centering
      \includegraphics[width=1.0\textwidth]{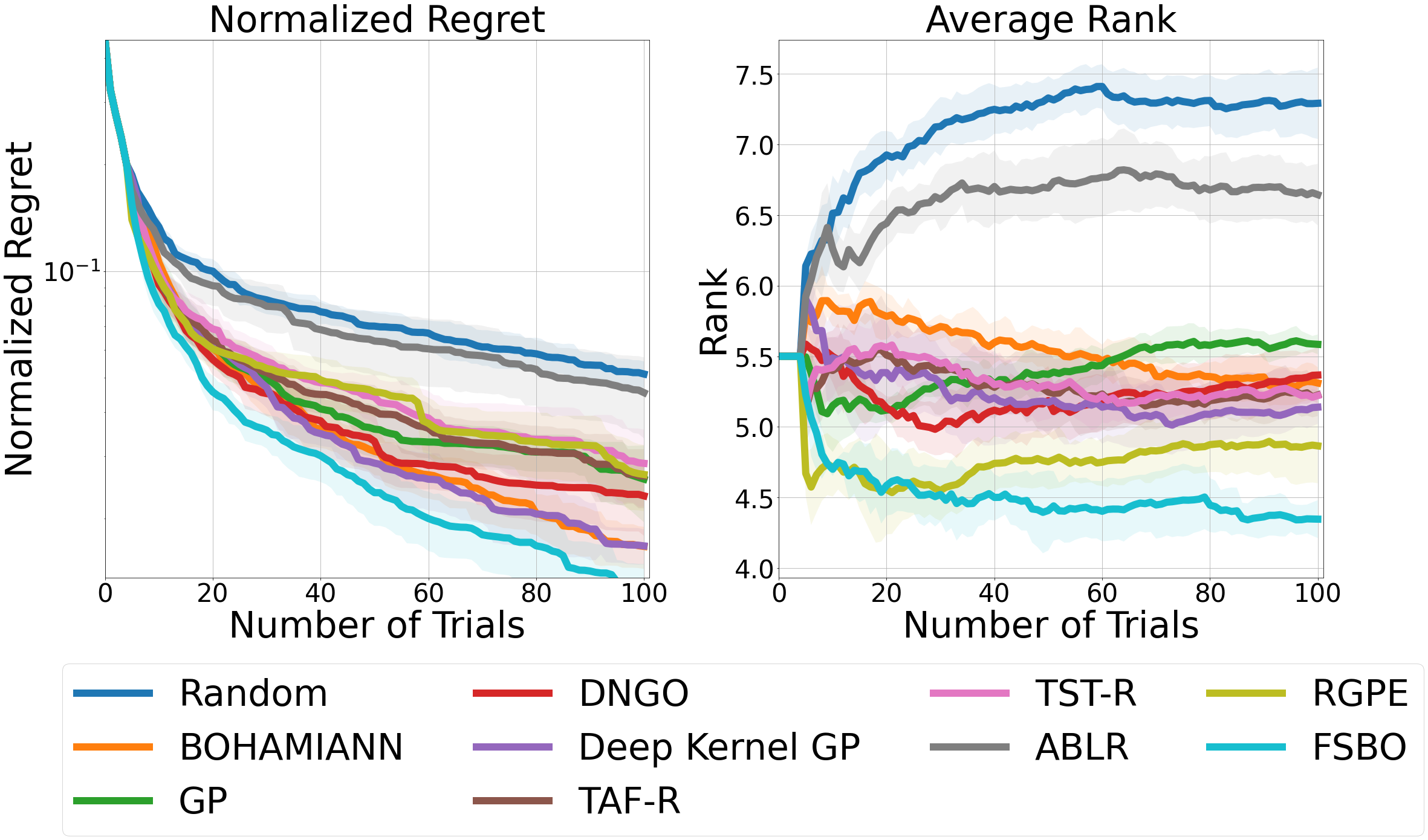}
\end{subfigure} 
\begin{subfigure}[b]{0.41\textwidth}
\includegraphics[width=1.0\textwidth]{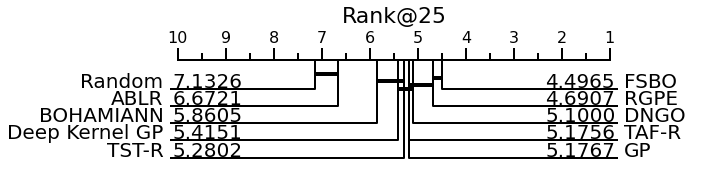}
\medskip
\includegraphics[width=1.0\textwidth]{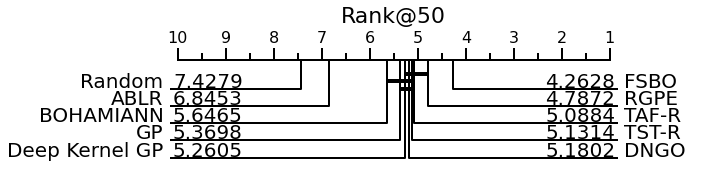}
\medskip
\includegraphics[width=1.0\textwidth]{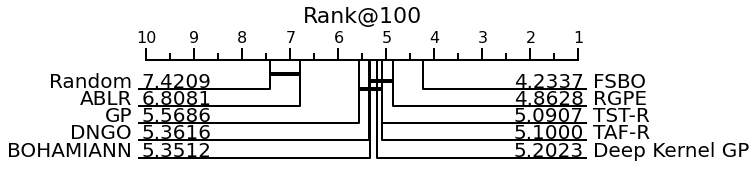}
\end{subfigure}
 \caption{Comparisons of normalized regret and mean ranks across all search spaces for the \textbf{transfer learning} and \textbf{non-transfer} HPO methods on \benchname{}-v3}
    \label{fig:averageall}
\end{figure}

We provide a cumulative comparison of both non-transfer and transfer black-box methods in Figure~\ref{fig:averageall}, for demonstrating the benefit of transfer learning in \benchname{}-v3. We see that the transfer methods (FSBO, RGPE, TST-R, TAF-R) achieve significantly better performances than the non-transfer techniques (GP, DNGO, BOHAMIANN, Deep Kernel GP). On the average rank plot and the associated Critical Difference diagrams, we notice that FSBO~\cite{wistuba2021few} achieves significantly better results than all baselines, followed by RGPE~\cite{feurer2018scalable}. A detailed comparison of the ranks per search-space is presented in the supplementary material. In particular, the direct gain of transfer learning can be observed by the dominance that FSBO has over \textit{Deep Kernel GP}, considering that both use exactly the same surrogate model and the same acquisition function. In comparison, the deep kernel parameters in FSBO are initialized from the solution of a meta-learning optimization conducted on the meta-train tasks of \benchname{}-v3 (transfer), while the parameters of \textit{Deep Kernel GP} are initialized randomly (no transfer).

\subsection{Validating the Continuous \benchname{} Benchmark}
\label{section:validating_continuous}
\begin{figure}[htb!]
\centering
\begin{subfigure}[b]{0.58\textwidth}
      \centering
      \includegraphics[width=0.45\textwidth]{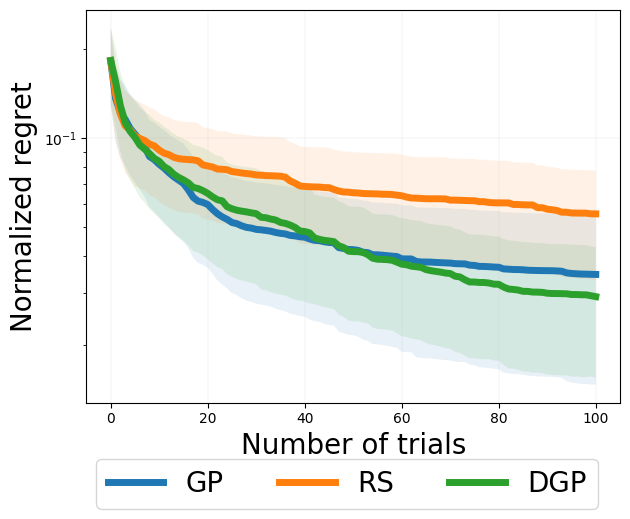}
    \includegraphics[width=0.45\textwidth]{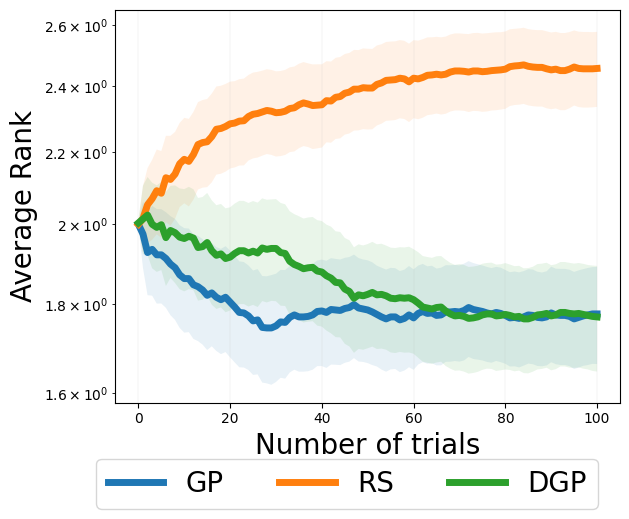}
\end{subfigure} 
\begin{subfigure}[b]{0.41\textwidth}
\includegraphics[width=1.0\textwidth]{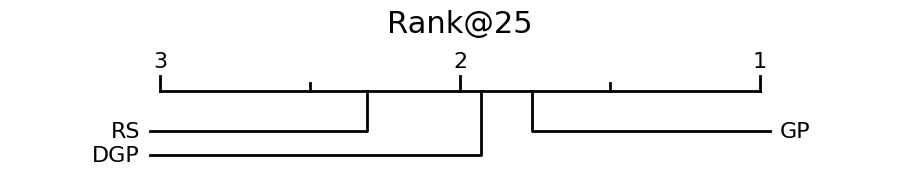}
\medskip
\includegraphics[width=1.0\textwidth]{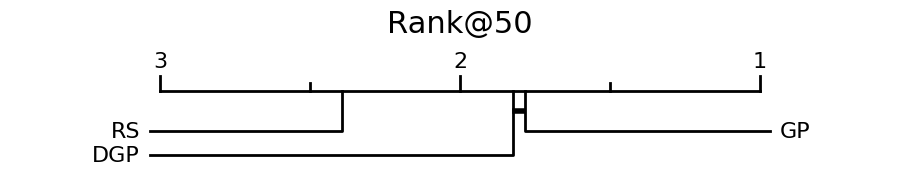}
\medskip
\includegraphics[width=1.0\textwidth]{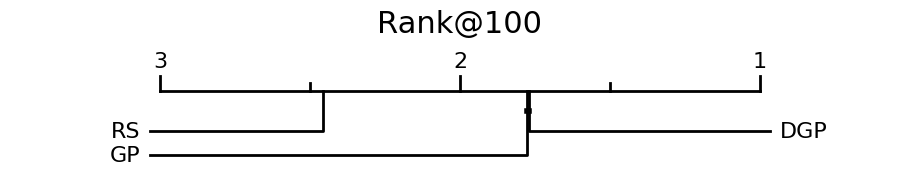}
\end{subfigure}
 \caption{\textbf{Aggregated} comparisons of normalized regret and mean ranks across all search spaces for three typical non-transfer HPO methods on the continuous search spaces of \benchname{}-v3}
    \label{fig:results_continuous}
\end{figure}

We further show that the surrogate-based continuous variant of HPO-B (Section~\ref{section:benchmark_continuous}) provides a benchmark where HPO methods achieve similar performances compared to the discrete HPO-B. We present the results of three typical non-transfer HPO methods (Gaussian Process (GP), Deep Kernel GP, and Random Search) on the continuous benchmark in Figure~\ref{fig:results_continuous}. The cumulative performance on the continuous surrogate tasks matches well with the performance of these methods on the discrete tasks (Figure~\ref{fig:averageablation}). In particular, we highlight a similar comparative trend of Deep Kernel GP being marginally better than GP after many trials but significantly superior to Random Search.

\section{Limitations of HPO-B}


A limitation of HPO-B is that it only covers black-box HPO tasks, instead of other HPO problems, such as grey-box/multi-fidelity HPO, online HPO, or pipeline optimization for AutoML libraries. In addition, HPO-B is restricted by the nature of search spaces found in OpenML, which contains evaluations for well-established machine learning algorithms for tabular data, but lacks state-of-the-art deep learning methods, or tasks involving feature-rich data modalities (image, audio, text, etc.). An additional limitation is the structured bias and noise produced by relying on a surrogate for constructing continuous search spaces. However, it has been found that tree-based models are able to model the performance of several machine learning algorithms and produce surrogates that resemble real-world problems~\cite{Eggensperger2015Efficient}. Other sources of bias and noise might come from the user-oriented data generation process for the evaluation on discrete search spaces, which might potentially incur in wrong values or hyperparameters within ranges reflecting prior knowledge or typical human choices. These risks can be reduced by benchmarking on a large number of search spaces, as we suggested throughout the paper. 

\section{Conclusions}

Recent HPO and transfer-learning HPO papers inconsistently use different meta-datasets, arbitrary train/validation/test splits, as well as ad-hoc preprocessing, which makes it hard to reproduce the published results. To resolve this bottleneck, we propose HPO-B, a novel benchmark based on the OpenML repository, that contains meta-datasets from 176 search spaces, 196 datasets, and a total of 6.4 million evaluations.  For promoting reproducibility at a \textit{level playing field} we also provide initial configuration seeds, as well as predefined training, validation and testing splits. Our benchmark contains pre-processed meta-datasets and a clear set of HPO tasks and exact splits, therefore, it enables future benchmark results to be directly comparable. We believe our benchmark has the potential to become the \textit{de facto} standard for experimentation in the realm of black-box HPO.


\section*{Acknowledgements}
The research of Hadi S. Jomaa is co-funded by the industry project \href{https://www.ismll.uni-hildesheim.de/projekte/ecosphere_en.html} {"IIP-Ecosphere: Next Level Ecosphere for Intelligent Industrial Production"}. Prof. Grabocka is also thankful to the Eva Mayr-Stihl Foundation for their generous research grant, as well as to the Ministry of Science, Research and the Arts of the German state of Baden-W\"urttemberg, and to the BrainLinks-BrainTools Excellence Cluster for funding his professorship. In addition, we thank Arlind Kadra for his assistance in interfacing with the OpenML Python package.

\bibliographystyle{plain}
\bibliography{bibliography}
\newpage
\section*{Checklist}

The checklist follows the references.  Please
read the checklist guidelines carefully for information on how to answer these
questions.  For each question, change the default \answerTODO{} to \answerYes{},
\answerNo{}, or \answerNA{}.  You are strongly encouraged to include a {\bf
justification to your answer}, either by referencing the appropriate section of
your paper or providing a brief inline description.  For example:
\begin{itemize}
  \item Did you include the license to the code and datasets? \answerYes{See Section}
  \item Did you include the license to the code and datasets? \answerNo{The code and the data are proprietary.}
  \item Did you include the license to the code and datasets? \answerNA{}
\end{itemize}
Please do not modify the questions and only use the provided macros for your
answers.  Note that the Checklist section does not count towards the page
limit.  In your paper, please delete this instructions block and only keep the
Checklist section heading above along with the questions/answers below.

\begin{enumerate}

\item For all authors...
\begin{enumerate}
  \item Do the main claims made in the abstract and introduction accurately reflect the paper's contributions and scope?
    \answerYes{}
  \item Did you describe the limitations of your work?
    \answerYes{}
  \item Did you discuss any potential negative social impacts of your work?
    \answerNA{}
  \item Have you read the ethics review guidelines and ensured that your paper conforms to them?
    \answerYes{}
\end{enumerate}

\item If you are including theoretical results...
\begin{enumerate}
  \item Did you state the full set of assumptions of all theoretical results?
    \answerNA{}
	\item Did you include complete proofs of all theoretical results?
    \answerNA{}
\end{enumerate}

\item If you ran experiments (e.g. for benchmarks)...
\begin{enumerate}
  \item Did you include the code, data, and instructions needed to reproduce the main experimental results (either in the supplemental material or as a URL)?
    \answerYes{See our repository link in the introduction. }
  \item Did you specify all the training details (e.g., data splits, hyperparameters, how they were chosen)?
    \answerNo{We used pre-defined configurations from previous work.}
	\item Did you report error bars (e.g., concerning the random seed after running experiments multiple times)?
    \answerYes{}
	\item Did you include the total amount of computing and the type of resources used (e.g., type of GPUs, internal cluster, or cloud provider)?
    \answerNo{}
\end{enumerate}

\item If you are using existing assets (e.g., code, data, models) or curating/releasing new assets...
\begin{enumerate}
  \item If your work uses existing assets, did you cite the creators?
    \answerYes{}
  \item Did you mention the license of the assets?
    \answerNo{They are included in the cited publications.}
  \item Did you include any new assets either in the supplemental material or as a URL?
    \answerYes{}
  \item Did you discuss whether and how consent was obtained from people whose data you're using/curating?
    \answerNo{We are using open-sourced assets.}
  \item Did you discuss whether the data you are using/curating contains personally identifiable information or offensive content?
    \answerNA{}
\end{enumerate}

\item If you used crowdsourcing or researched with human subjects...
\begin{enumerate}
  \item Did you include the full text of instructions given to participants and screenshots, if applicable?
    \answerNA{}
  \item Did you describe any potential participant risks, with links to Institutional Review Board (IRB) approvals, if applicable?
    \answerNA{}
  \item Did you include the estimated hourly wage paid to participants and the total amount spent on participant compensation?
    \answerNA{}
\end{enumerate}

\end{enumerate}

\newpage
\appendix

\section{Datasheet}
\subsection{Motivation}
\begin{itemize}
    \item \textbf{For what purpose was the dataset created?} In order to benchmark easily new methods on HPO and transfer HPO.
    \item \textbf{Who create the dataset and on behalf of which entity?} The authors of the main paper work together on behalf of their respective institution.
    \item \textbf{Who funded the creation of the dataset?} University of Freiburg, University of Hildesheim and Amazon.
\end{itemize}
\subsection{Composition}
\begin{itemize}
    \item \textbf{What do the instances that comprises the dataset represent?} They represent evaluations of hyperparameter configurations on diferent tasks.
    \item \textbf{How many instances are there in total?} Around 6.4 million hyperparameter evaluations.
    \item \textbf{Does the dataset contain all possible instances or is it a sample of instances from a larger set?} No, they correspond only to evaluations of supervised classifiers and queried from OpenML platform.
    \item \textbf{What data does each instance consist of?} Each instance consist of a hyperparameter configuration, with their respective values and response. 
    \item \textbf{Is there a label or target associated with each instance?} Yes, the associated accuracy (validation accuracy).
    \item \textbf{Is any information missing from individual instances?} No. The missing values have been imputed.
    \item \textbf{Are relationships between individual instances made explicit?} No. 
    \item \textbf{Are there recommended data splits?} Yes, we recommend three splits for meta-train, meta-validation and meta-test.
    \item \textbf{Are there any errors, sources of noise or redundancies in the dataset?} Yes, the evaluations rely on third parties that may have commited mistakes in reporting the response or hyperparameter configuration values.
    \item \textbf{Is the dataset self-contained or does it link to or otherwise rely on external resources?} For the creation, it relied on data from OpenML. 
    \item \textbf{Does the dataset contain data that might be considered confidential?} No.
    \item \textbf{Does the dataset contain data that, if viewed directly, might be offensive, insulting, threatening or might otherwise cause anxiety?} No.
    \item \textbf{Does the dataset relate to people?} No.
    \item \textbf{Does the dataset identify any sub-populations?} No.
    \item \textbf{Is it possible to identify individuals, either directly or indirectly?} No.
    \item \textbf{Does the dataset contain data that might be considered sensitive in any way?} No.
\end{itemize}

\subsection{Collection process}

\begin{itemize}
    \item \textbf{How was the data associated with each instance acquired?} We queried the existing runs (evaluations) with tag \texttt{Verified\_Supervised\_Classification} from OpenML.
    \item \textbf{What mechanisms or procedures were used to collect the data?} We used the Python API for fetching the data.
    \item \textbf{Who was involved in the data collection process?}  Only the authors of this paper.
    \item \textbf{Over what time frame was the data collected?} The existing runs until April 15, 2021.
    \item \textbf{Were any ethical review process conducted?} No.
    \item \textbf{Does the dataset relate to people?} No.
    \item \textbf{Did you collect the data from the individuals in question, or obtained it via third parties or other sources?} It was obtained from OpenML website.
    \item \textbf{Was any preprocessing/cleaning/labeling of the data done?} Yes, it is explain in section \ref{sec:compandprep}.
    \item \textbf{Was the "raw" data saved in addition to the preprocessed/cleaned/labeled data?} No, but the raw data can be accessed through OpenML platform.
    \item \textbf{Is the software used to preprocess/clean/label the instances available?} No.
\end{itemize}

\subsection{Uses}
\begin{itemize}
    \item \textbf{Has the dataset been used for any tasks already?} Yes, we provide examples in section \ref{sec:results}.
    \item \textbf{Is there a repository that links to any or all papers or systems that use the dataset?} Yes, it is available in  \url{https://github.com/releaunifreiburg/HPO-B}.
    \item \textbf{What (other) tasks could the dataset be used for?} HPO and transfer HPO. 
    \item \textbf{Is there anything about the composition of the dataset or the way it was collected and preprocessed/cleaned/labeled that might impact future uses?} No.
    \item \textbf{Are there tasks for which the dataset should not be used?} No.
\end{itemize}

\subsection{Distribution}
\begin{itemize}
    \item \textbf{Will the dataset be distributed to third parties outside of the entity?}  Yes, the dataset is publicly available.
    \item \textbf{How will the dataset be distributed?} Through a link hosted in our repository \url{https://github.com/releaunifreiburg/HPO-B}.
    \item \textbf{When will the dataset be distributed?} From June 8. 2021.
    \item \textbf{Will the dataset be distributed under a copyright or other intellectual property license?} Yes, license CC-BY.
    \item \textbf{Have any third parties imposed IP-based or other restrictions on the data associated with the instances?} No.
    \item \textbf{Do any export controls or other regulatory restrictions apply to the dataset or to individual instances?} No.
\end{itemize}
\subsection{Maintenance}
\begin{itemize}
    \item \textbf{Who will be supporting/hosting/maintaining the dataset?} RELEA group from the University of Freiburg.
    \item \textbf{How can the owner/curator/manager of the dataset be contacted?} Questions can be sent to \texttt{pineda@cs.uni-freiburg.de} or submit an issue to the Github repository \url{https://github.com/releaunifreiburg/HPO-B}.
    \item \textbf{Is there an erratum?} No.
    \item \textbf{Will the dataset be updated?} No.
    \item \textbf{Will older versions of the dataset continue to be supported/hosted/maintained?} As new runs (evaluations) might be available in OpenML, we consider to update it periodically.
    \item \textbf{If other want to extend/augment/build on/contributed to the dataset, is there a mechanism for them to do so?} We might provide this possibility in the future.
\end{itemize}

\section{License}

The meta-dataset is provided under a license CC-BY.

\section{Accessibility}

The link to access the meta-dataset and recommendations on how to consume it are given in the following git-hub repository: \url{https://github.com/releaunifreiburg/HPO-B}. 

\section{Maintenance}

We are planning to increase the meta-data set, as new search spaces with a considerable amount of evaluations are available on the platform. We may also add evaluations from third parties that agree to be part of the meta-dataset. For any suggestion or technical inquiry, we recommend to use the issue tracker of our repository.

\section{Additional Meta-dataset Statistics}
\label{sec:statistics}

\begin{figure}[ht]
\begin{subfigure}{.5\textwidth}
  \centering
  \includegraphics[width=.8\linewidth]{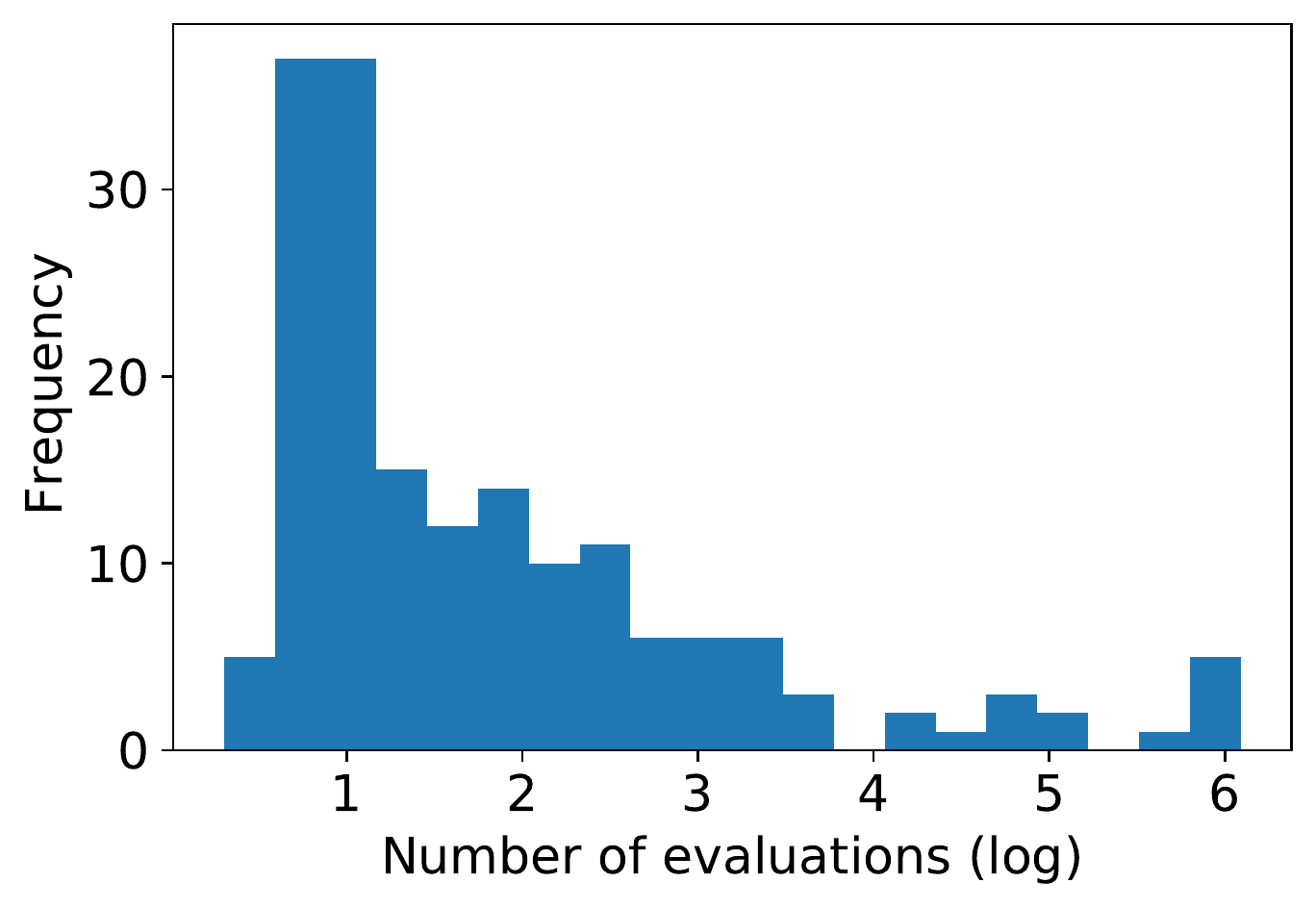}  
  \caption{Frequency of evaluations per search space.}
  \label{fig:sub-first}
\end{subfigure}
\begin{subfigure}{.5\textwidth}
  \centering
  \includegraphics[width=.8\linewidth]{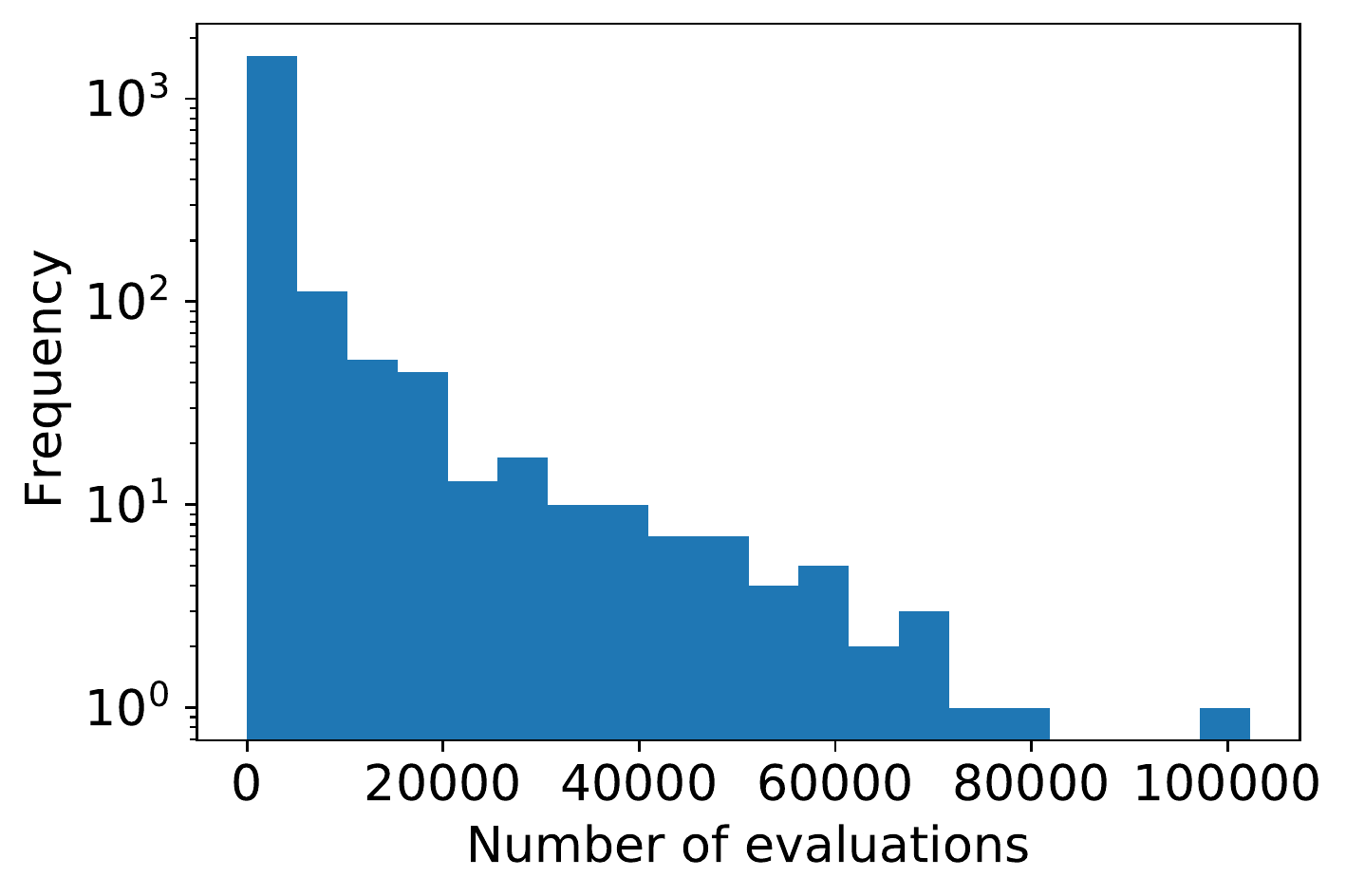}  
  \caption{Frequency of the number of evaluations per task.}
  \label{fig:sub-second}
\end{subfigure}
\begin{subfigure}{.5\textwidth}
  \centering
  \includegraphics[width=.8\linewidth]{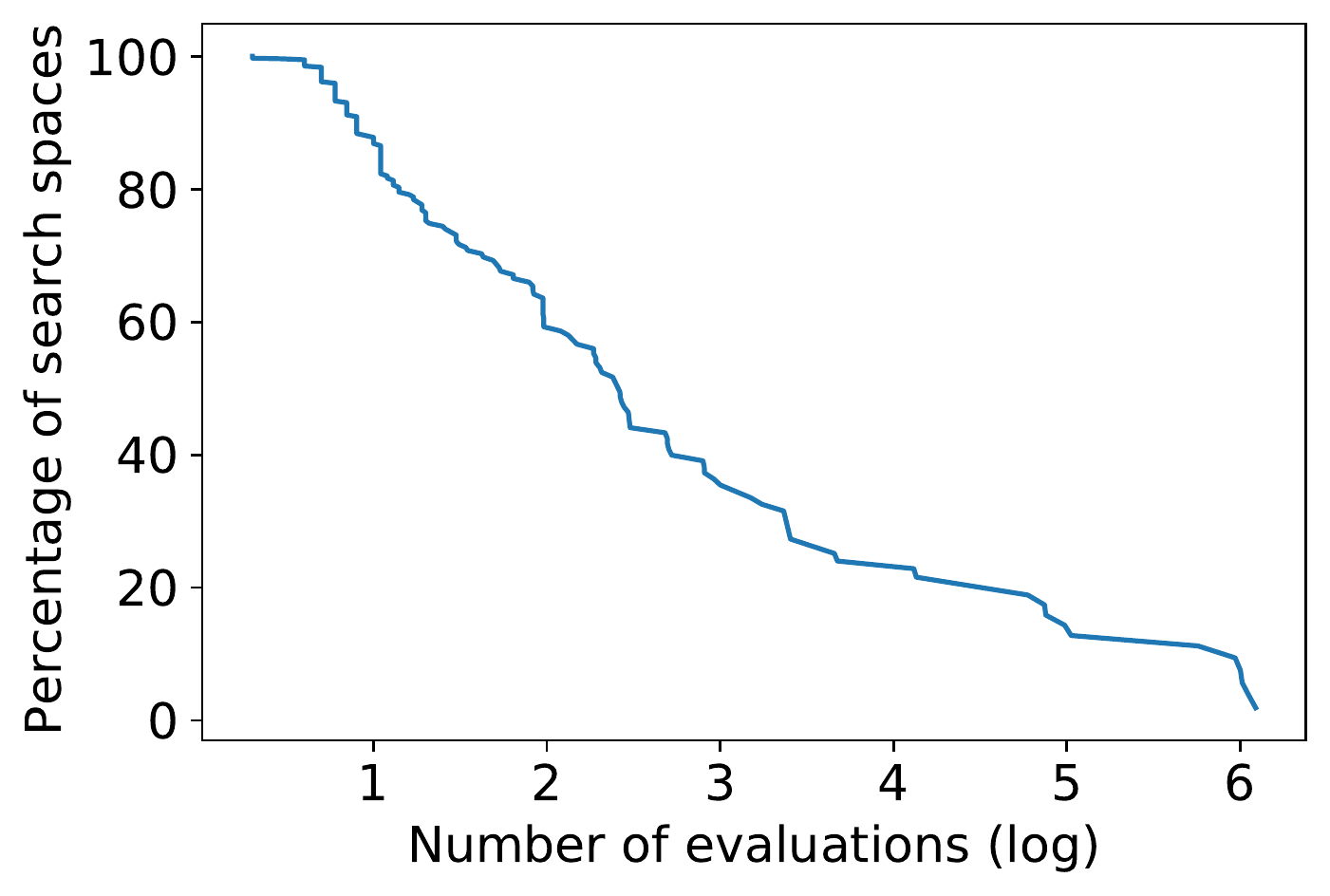}  
  \caption{Percentage of search spaces with a given number of evaluations.}
  \label{fig:sub-third}
\end{subfigure}
\begin{subfigure}{.5\textwidth}
  \centering
  \includegraphics[width=.8\linewidth]{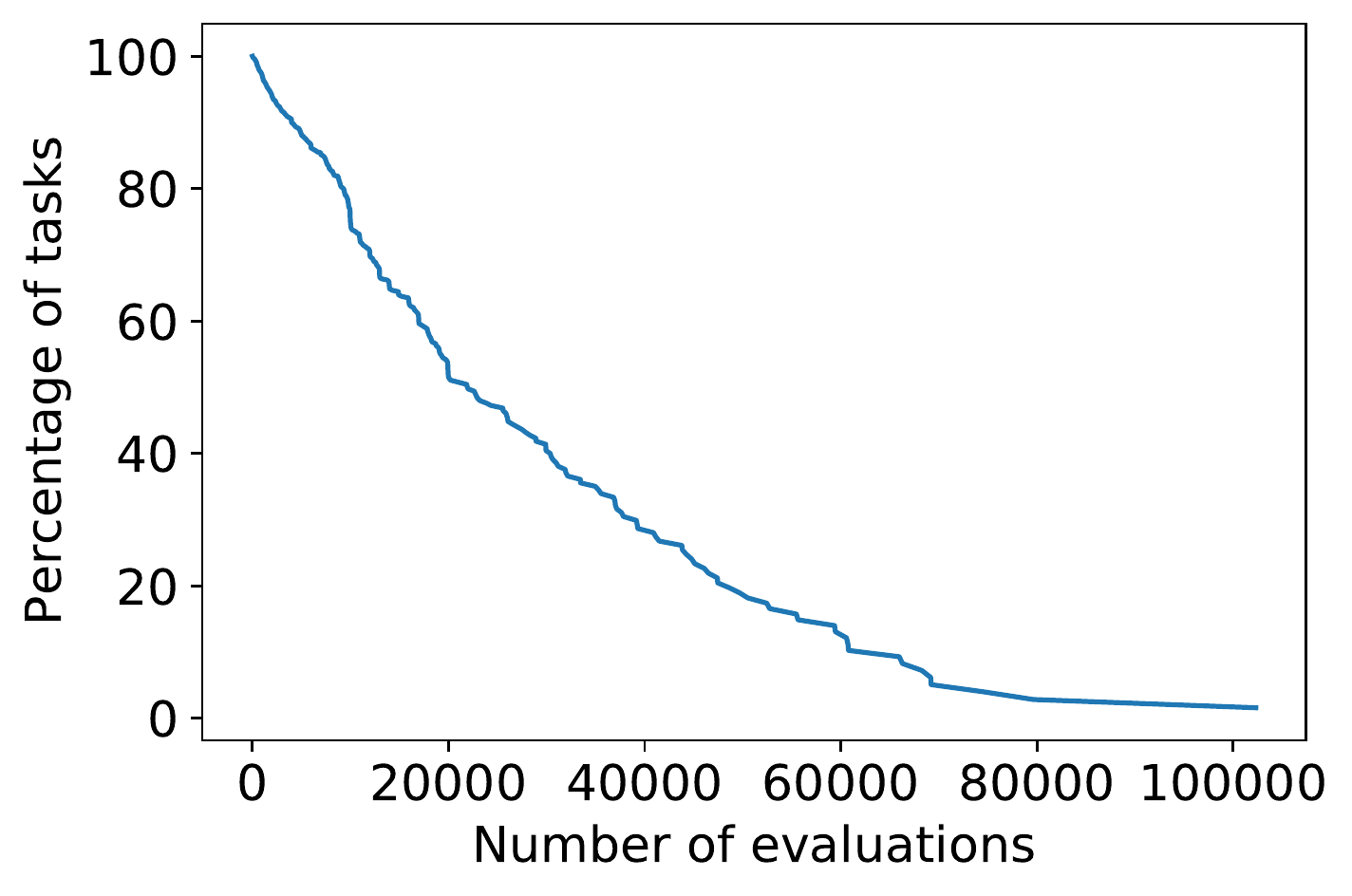}  
  \caption{Percentage of search spaces with a given number of evaluations.}
  \label{fig:sub-fourth}
\end{subfigure}
\caption{Descriptors for the number of evaluations.}
\label{fig:desc_number_of_evaluations}
\end{figure}

In this section, we present further descriptive statistics of the meta-dataset. The histogram in Figure~\ref{fig:sub-first} depicts how many search spaces (out of the 176) contain a specific number of evaluations. At the task level, the histogram shows a similar pattern in Figure~\ref{fig:sub-second}. The Figures ~\ref{fig:sub-third} and ~\ref{fig:sub-fourth} show respectively the percentage of search spaces and tasks with \textit{at least} a given number of evaluations. As can be seen, 50\% of the tasks have at least 20000 evaluations, and at least 30\% contain 40.000 evaluations (out of 1907 tasks). 

\begin{figure}[ht!]
    \centering
    \includegraphics[width=\textwidth]{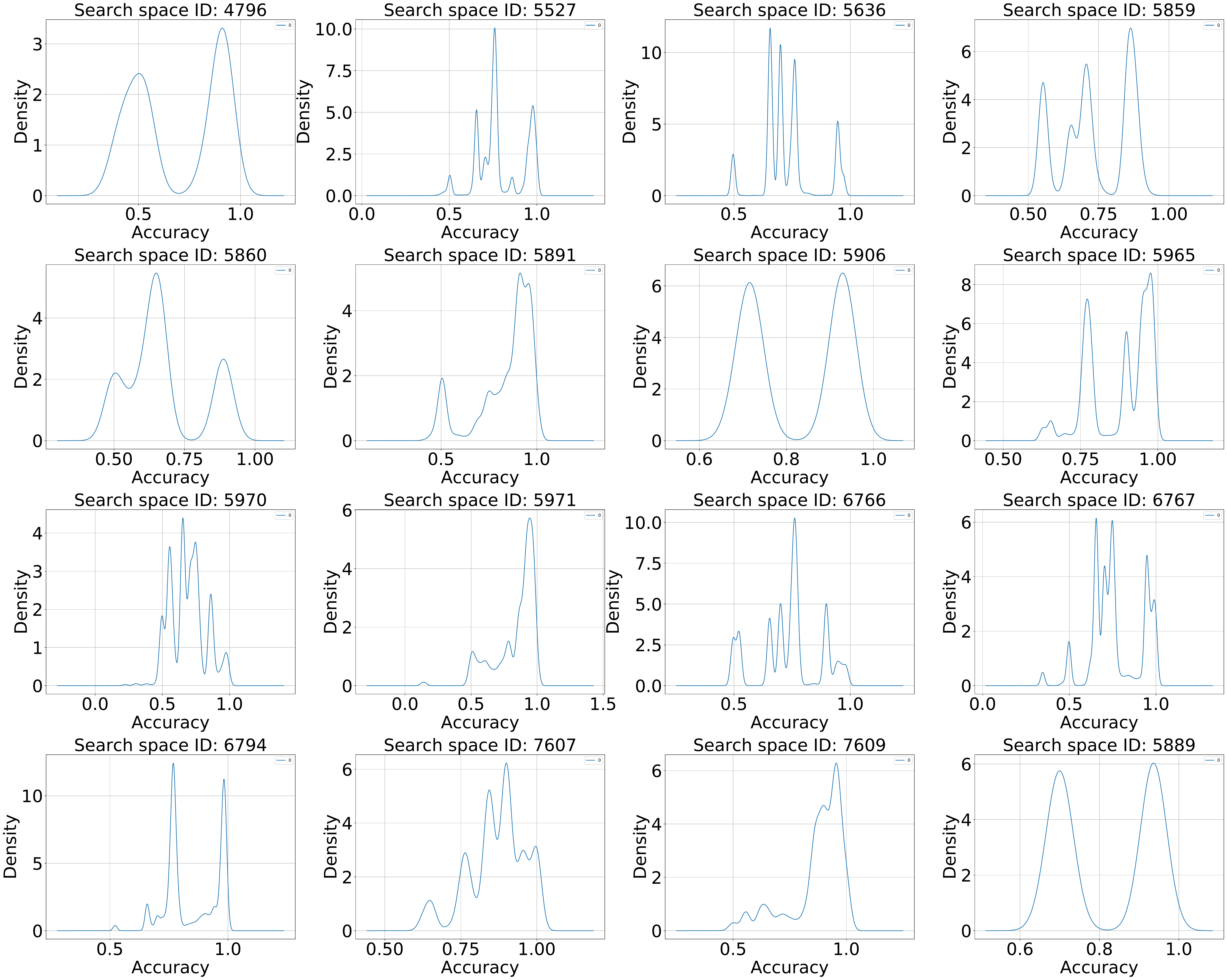}
    \caption{Density distribution of the response per search-space.}
    \label{fig:evaluations_kde}
\end{figure}

\begin{figure}[ht!]
    \centering
    \includegraphics[width=1.0\textwidth]{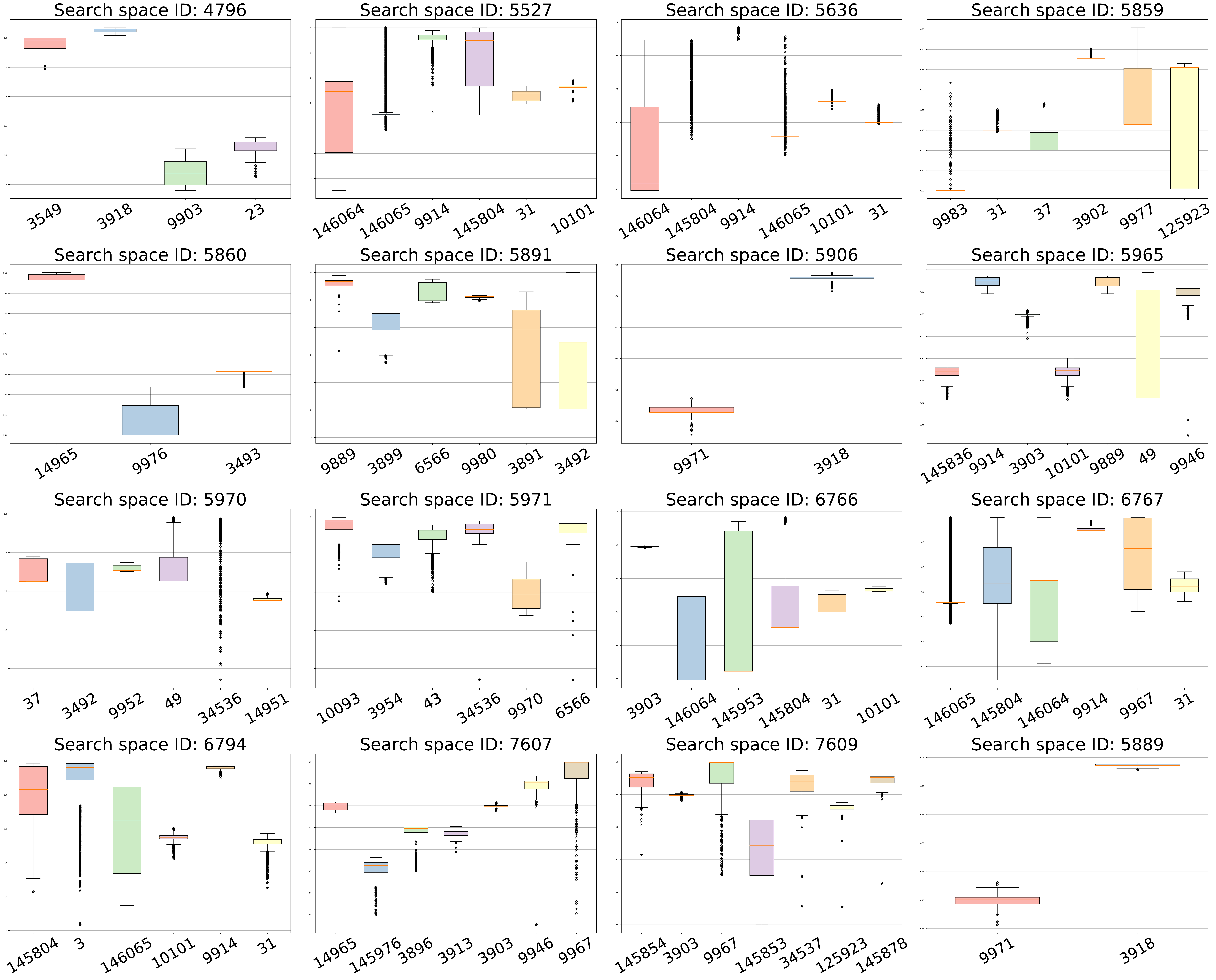}
    \caption{Distribution of the response (accuracy) per task in meta-test (Box-plot ranges are between 0 and 1).}
    \label{fig:response_per_task}
\end{figure}

On another discourse, the distribution of the response per search-space in the meta-test split can be observed in Figure~\ref{fig:evaluations_kde}. It is noticeable that the distributions are multi-modal, due to the fact that they correspond to the aggregated response of aggregated tasks. It shows how every task has its own scale. A more-detailed insight of the distribution per dataset is shown in the Figure \ref{fig:response_per_task}.

\section{Additional Results}
\label{section:additional_results}

\begin{figure}[ht!]
    \centering
    \includegraphics[width=\textwidth]{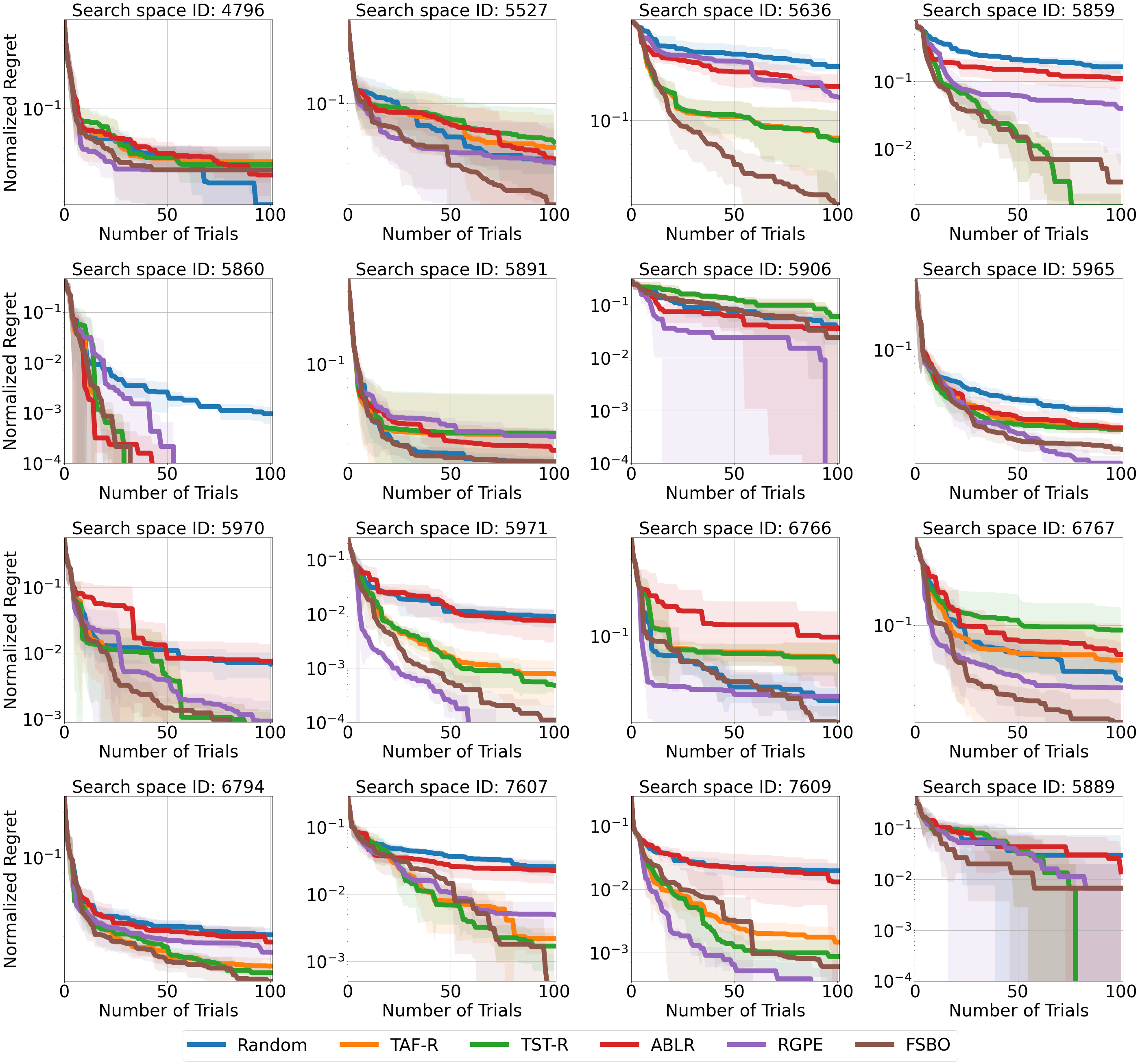}
    \caption{\textbf{Normalized regret} comparison of \textbf{transfer learning} HPO methods on \benchname{}-v3}
    \label{fig:regrettransfer}
\end{figure}

\begin{figure}[t]
    \centering
    \includegraphics[width=\textwidth]{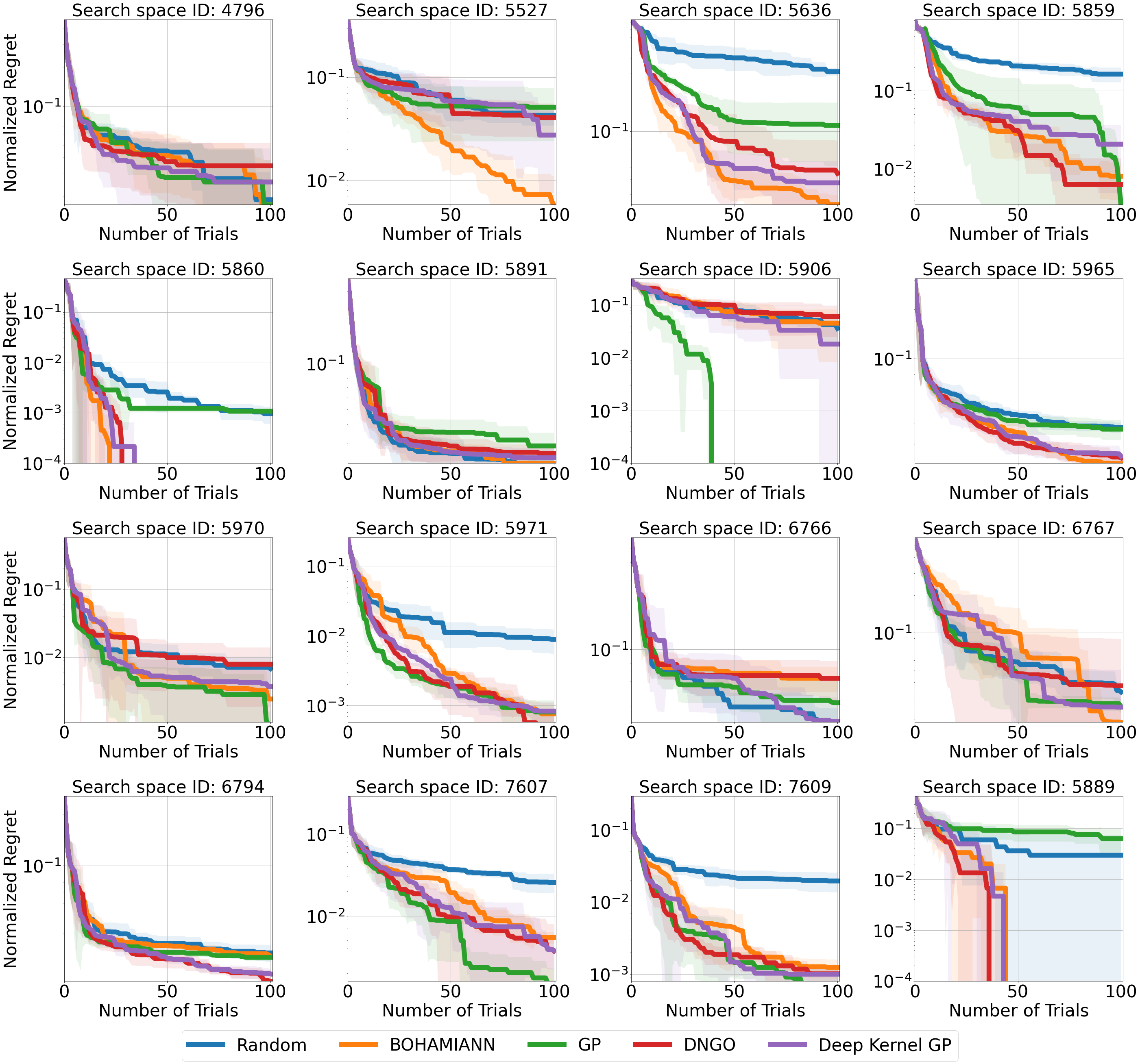}
    \caption{\textbf{Normalized regret} comparison of \textbf{non-transfer} black-box HPO methods on \benchname{}-v2}
    \label{fig:regretnontransfer}
\end{figure}

\begin{figure}[ht!]
    \centering
    \includegraphics[width=\textwidth]{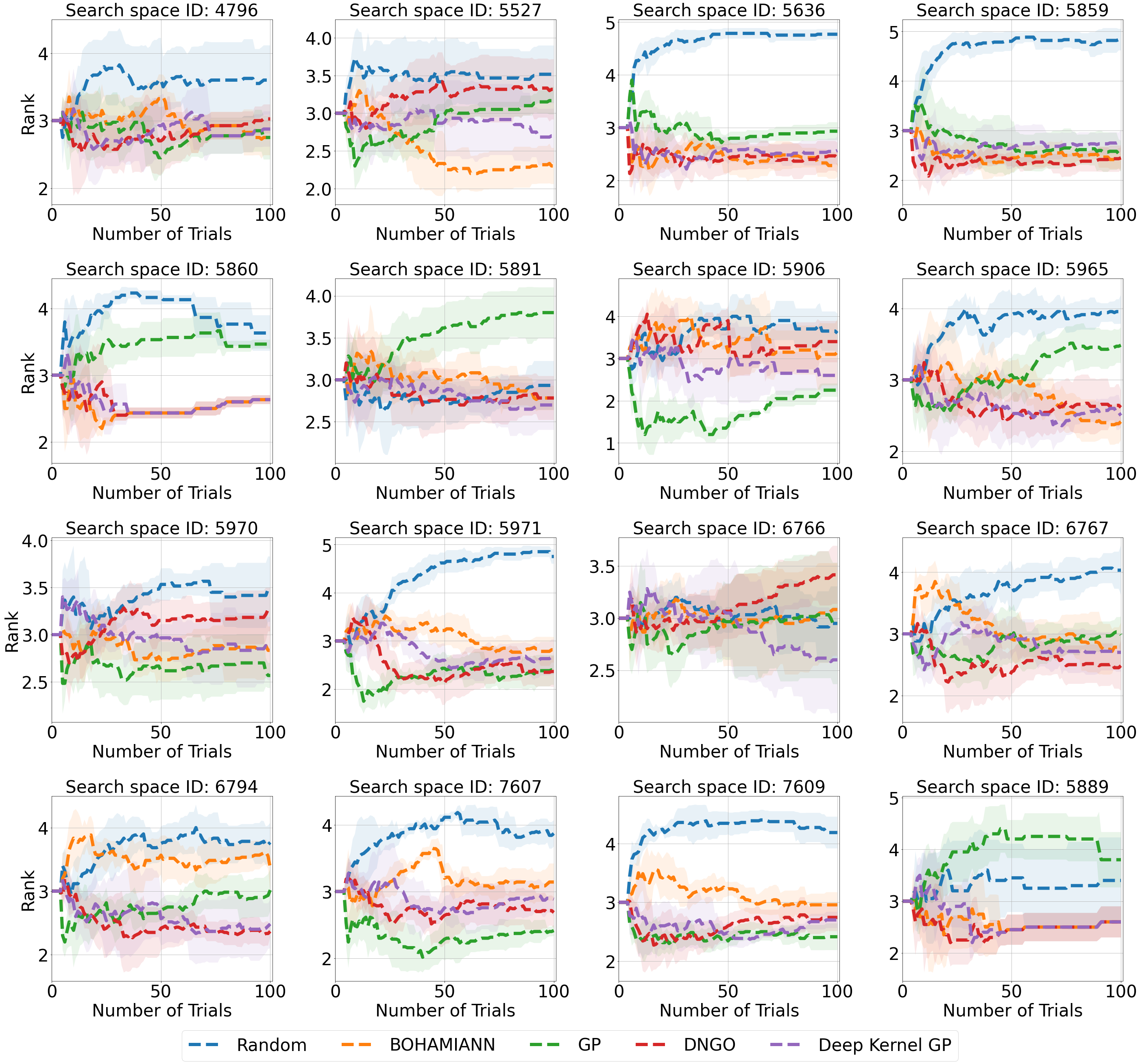}
    \caption{\textbf{Mean rank} comparison of \textbf{non-transfer} black-box HPO methods}
    \label{fig:ranknontransfer}
\end{figure}

\begin{figure}[htb!]
    \centering
    \includegraphics[width=\textwidth]{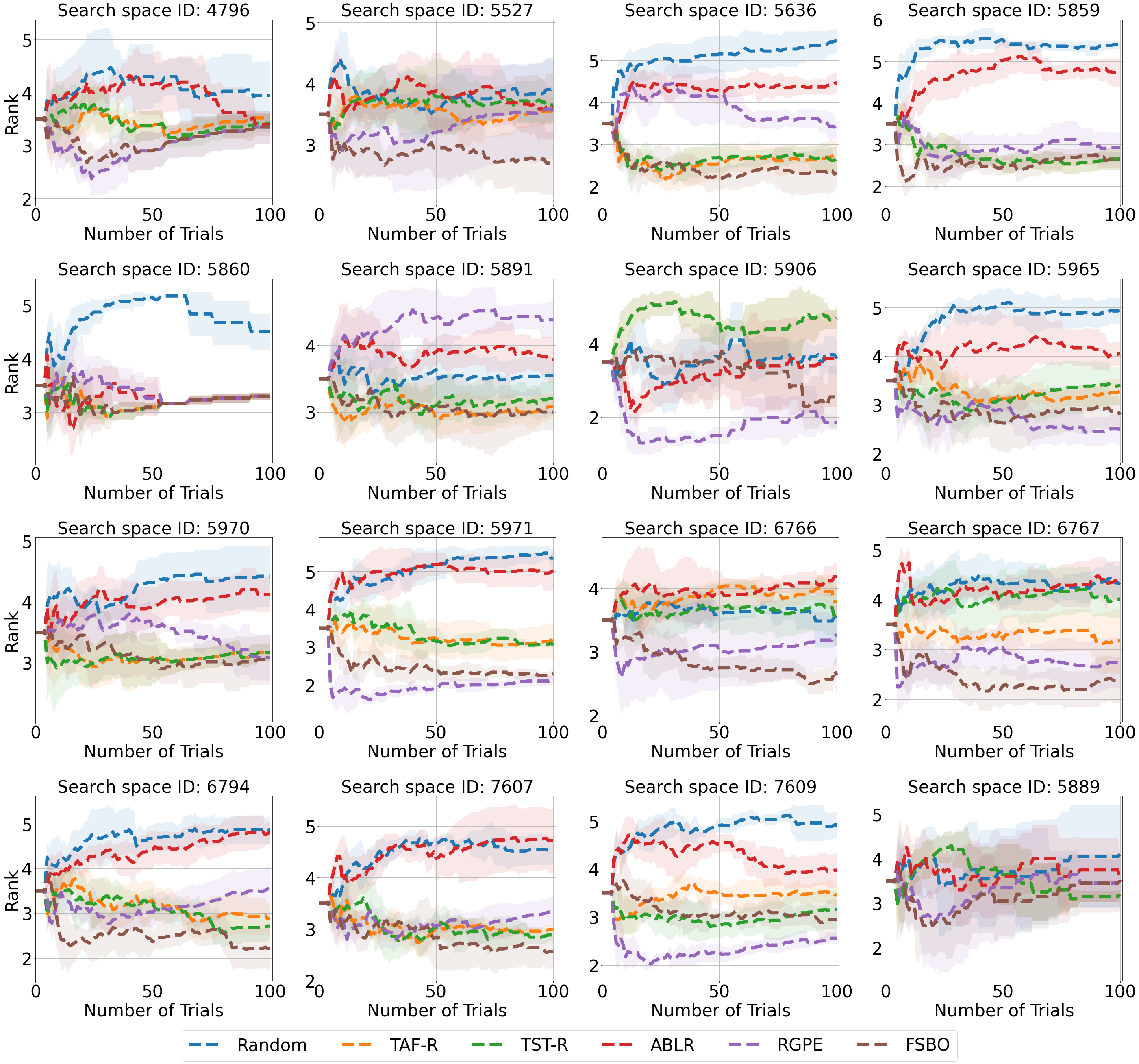}
    \caption{\textbf{Mean rank} comparison of \textbf{transfer learning} black-box HPO methods}
    \label{fig:ranktransfer}
\end{figure}

\begin{figure}[htb!]
    \centering
    \includegraphics[width=1.1\textwidth]{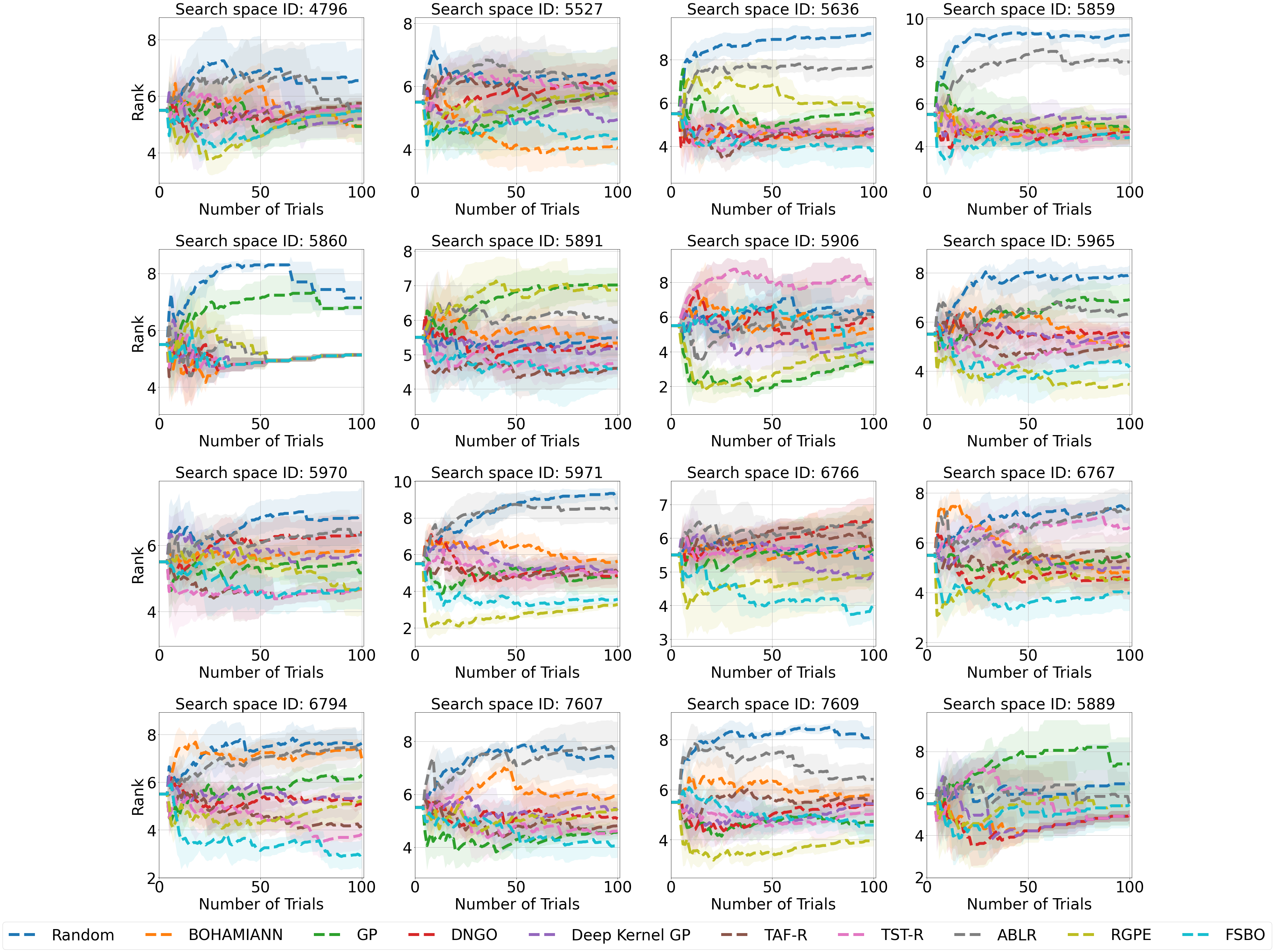}
    \caption{\textbf{Mean rank} comparison of \textbf{non-transfer} and \textbf{transfer learning} black-box HPO methods}
    \label{fig:ranktransfernontransfer}
\end{figure}

\begin{figure}[ht!]
    \centering
    \includegraphics[width=\textwidth]{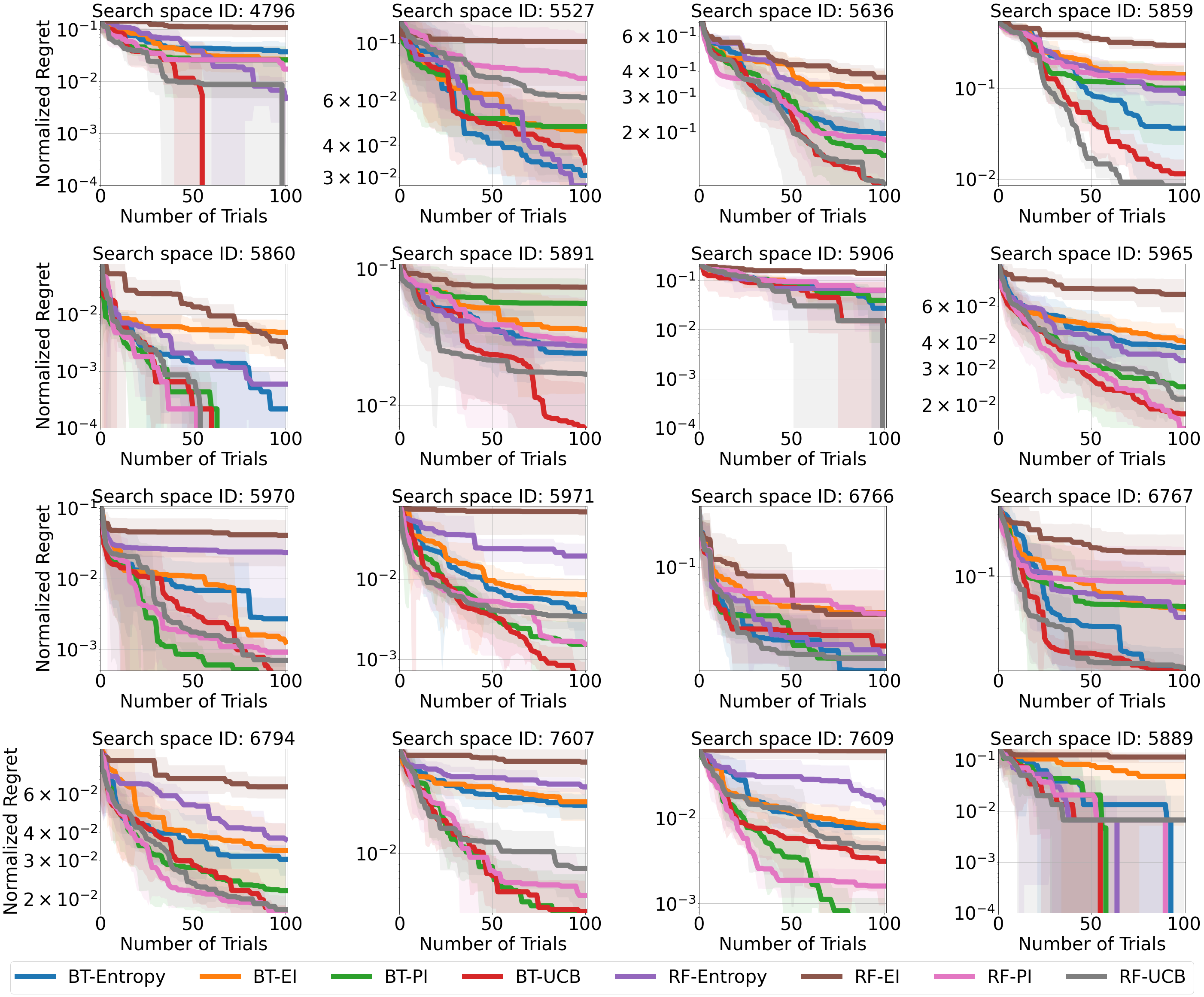}
    \caption{\textbf{Normalized regret} comparison of different acquisition functions: Expected Improvement (EI), Entropy (E) and Upper Confidence Bound (UCB) for Random Forest (RF) and Boosted Trees (BT) as surrogates.}
    \label{fig:regretacquisition}
\end{figure}

\begin{figure}[ht!]
    \centering
    \includegraphics[width=\textwidth]{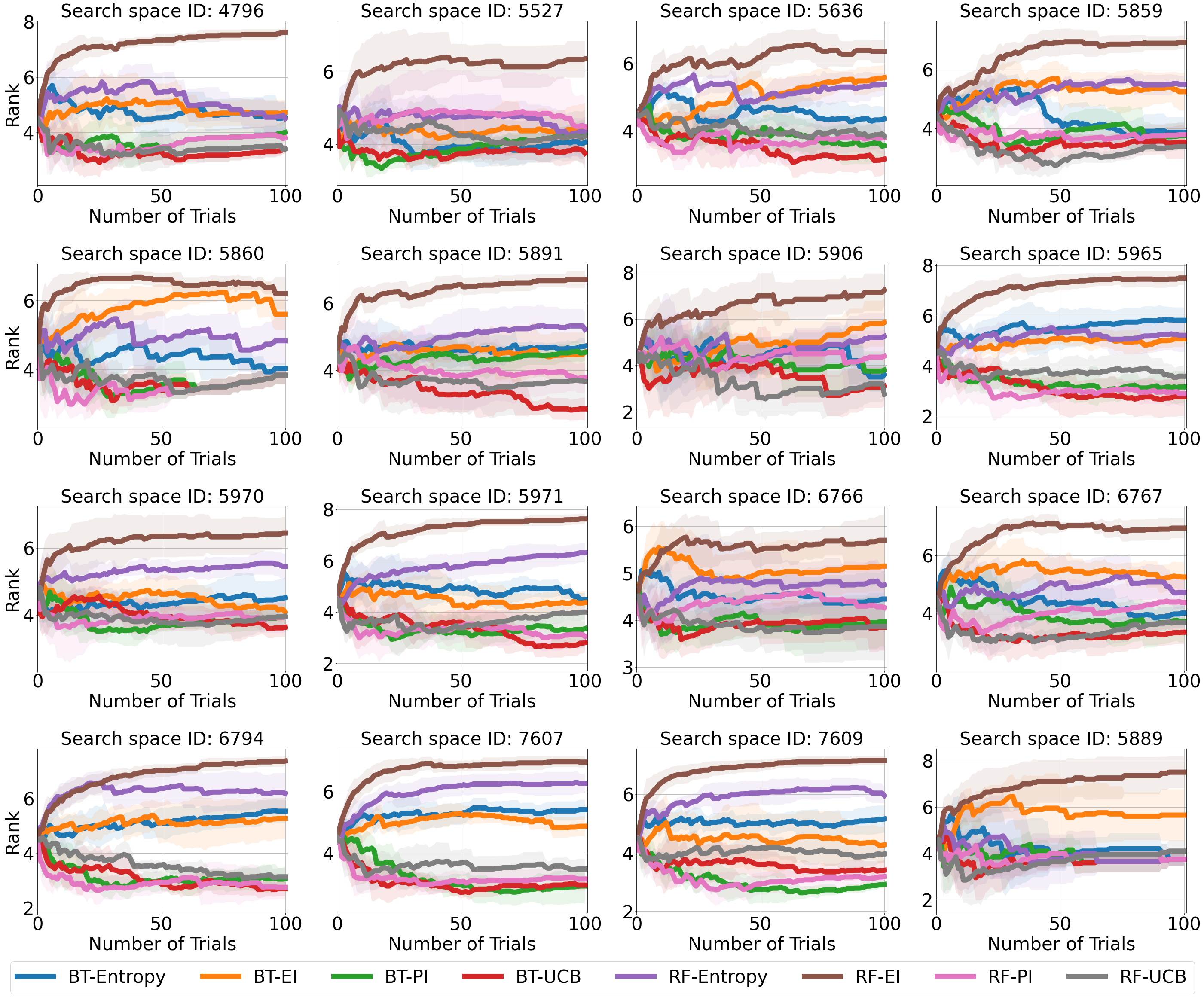}
    \caption{\textbf{Mean rank} comparison of different acquisition functions}
    \label{fig:rankacqusition}
\end{figure}

Figures \ref{fig:regrettransfer}, \ref{fig:regretnontransfer}, \ref{fig:ranktransfer}, \ref{fig:ranknontransfer}, and \ref{fig:rankacqusition} show the results of the average regret and rank, computed per task for transfer, non-transfer and acquisition methods.
Figure \ref{fig:ranktransfernontransfer} provides a comparison between transfer and non-transfer methods with respect to the average rank.
Moreover, Figure \ref{fig:regretacquisition} presents detailed results for the average rank with different acquisition functions, which summarized in the main paper. 

\section{Further Data-Preprocessing Details}
\label{section:further_preprocessing}

We explain in more details the data-extraction and pre-processing details for the benchmark. Additional information can also be obtained from the codebase in our repository. In the following explanation, a flow refers to a specific algorithm or search space.

\begin{itemize}
    \item We list all the existing flows under the tag \textit{Supervised Verified Classification}. Once we get the flows, we query all the IDs of the runs associated with this flows. As this process may overcharge the server, we perform it in batches.
    \item Subsequently, we perform a selection of flows and datasets by considering only those runs ID from dataset-flow instances (tasks) that contain more than 5 runs in total. This aims to decrease the number of runs from noisy small tasks.
    \item We query the actual runs. As this process may overload the servers, we perform it in batches. From all the information returned by the API in a Python-like dictionary, we just keep the keys \textit{run\_id, task\_id, flow\_name, accuracy} and \textit{parameter settings}. The last one refers to the specific hyperparameter configuration.
    \item As the hyperparameter configurations come as strings, we recast the values to either \textit{string, float} or \textit{integer}.
    \item We eliminate repeated evaluations and filter out tasks with only one evaluation.
    \item We group all the evaluations from the same flow but different datasets into a single data-frame. This allows having a single view of all the hyperparameter configurations used across datasets for the specific flow.
    \item We one-hot encode the categorical hyperparameters. Afterward, as some hyperparameter values may be missing for some datasets within a flow, we input missing values by setting them to zero. After imputation, we create an additional indicator variable per hyperparameter so that it is 1 if the value was imputed for the configuration and zero otherwise. 
    \item We get rid of hyperparameters which only have one value among the whole flow, as they are not a differential feature useful for assessing the performance.
    \item We apply the logarithmic transformation to a manually selected group of hyperparameters. 
    \item Finally, we normalize the values per hyperparameter across the whole flow, so that they vary between 0 and 1. We do not scale the response.
    
\end{itemize}

\begin{table}[ht]
\caption{Hyperparameters of the search spaces \benchname{}-v2/-v3}
\begin{tabular}{p{0.10\linewidth} p{0.20\linewidth}  p{0.6\linewidth}}
\hline
\textbf{Search Space Id} & \textbf{Search Space Name}       & \textbf{List of hyperparameters within the Search-space}                                                                                                                                                                                                       \\ \hline
4796                     & mlr.classif.rpart .preproc(16)    & minsplit, minbucket, cp                                                                                                                                                                                                                                        \\ \hline
5527                     & mlr.classif.svm (6)              & cost, gamma, gamma.na, degree, degree.na, kernel.ohe.na, kernel.ohe.linear, kernel.ohe.polynomial                                                                                                                                                              \\ \hline
5636                     & mlr.classif.rpart (29)   & minsplit, minsplit.na, minbucket, cp, maxdepth, maxdepth.na                                                                                                                                                                                                    \\ \hline
5859                     & mlr.classif.rpart (31)  & minsplit, minsplit.na, minbucket, cp, maxdepth, maxdepth.na                                                                                                                                                                                                    \\ \hline
5860                     & mlr.classif.glmnet (4)           & alpha, lambda                                                                                                                                                                                                                                                  \\ \hline
5891                     & mlr.classif.svm (7)              & cost, gamma, gamma.na, degree, degree.na, kernel.ohe.na, kernel.ohe.linear, kernel.ohe.polynomial                                                                                                                                                            \\ \hline
5906                     & mlr.classif.xgboost (4) & eta, max\_depth, max\_depth.na, min\_child\_weight, min\_child\_weight.na, subsample, colsample\_bytree, colsample\_bytree.na, colsample\_bylevel, colsample\_bylevel.na, lambda, alpha, nrounds, nrounds.na, booster.ohe.na, booster.ohe.gblinear                       \\ \hline
5965                     & mlr.classif.ranger(9)   & num.trees, num.trees.na, mtry, sample.fraction, min.node.size, min.node.size.na, replace.ohe.FALSE, replace.ohe.na, respect.unordered.factors.ohe.INVALID, respect.unordered.factors.ohe.TRUE                                                                  \\ \hline
5970                     & mlr.classif.glmnet(5)   & alpha, lambda                                                                                                                                                                                                                                                  \\ \hline
5971                     & mlr.classif.xgboost (6) & eta, max\_depth, max\_depth.na, min\_child\_weight, min\_child\_weight.na, subsample, colsample\_bytree, colsample\_bytree.na, colsample\_bylevel, colsample\_bylevel.na, lambda, alpha, nrounds, nrounds.na, booster.ohe.na, booster.ohe.gblinear                       \\ \hline
6766                     & mlr.classif.glmnet (11)          & alpha, lambda                                                                                                                                                                                                                                                  \\ \hline
6767                     & mlr.classif.xgboost (9)          & eta, subsample, lambda, alpha, nthread, nthread.na, nrounds, nrounds.na, max\_depth, max\_depth.na, min\_child\_weight, min\_child\_weight.na, colsample\_bytree, colsample\_bytree.na, colsample\_bylevel, colsample\_bylevel.na, booster.ohe.na, booster.ohe.gblinear \\ \hline
6794                     & mlr.classif.ranger(13)           & num.trees, num.trees.na, mtry, sample.fraction, min.node.size, min.node.size.na, replace.ohe.FALSE, replace.ohe.na, respect.unordered.factors.ohe.na respect.unordered.factors.ohe.TRUE                                                                        \\ \hline
7607                     & mlr.classif.ranger(15)           & num.trees, num.trees.na, mtry, min.node.size, sample.fraction, respect.unordered.factors.ohe.na, respect.unordered.factors.ohe.TRUE, replace.ohe.FALSE, replace.ohe.na                                                                                         \\ \hline
7609                     & mlr.classif.ranger(16)           & num.trees, num.trees.na, mtry, min.node.size, sample.fraction, respect.unordered.factors.ohe.na, respect.unordered.factors.ohe.TRUE, replace.ohe.FALSE, replace.ohe.na                                                                                         \\ \hline
5889                     & mlr.classif.ranger(7)            & num.trees, num.trees.na, mtry, sample.fraction, replace.ohe.FALSE, replace.ohe.na                                                                                                                                                                              \\ \hline
\end{tabular}
\end{table}

\begin{table}[ht]
\caption{All the search-spaces contained in \benchname{}-v1}
\begin{tabular}{p{0.10\linewidth} p{0.10\linewidth}  p{0.1\linewidth}  p{0.1\linewidth}  p{0.5\linewidth} }
\hline
\textbf{Search Space ID} & \textbf{Eval.} & \textbf{Datasets} & \textbf{Dim.} & \textbf{Search Space Name}                                                                \\ \hline
151                      & 25                       & 5                     & 1                       & weka.LMT                                                                                  \\ \hline
5886                     & 295                      & 33                    & 9                       & weka.J48                                                                                  \\ \hline
5918                     & 192             & 32                    & 13                      & weka.IBk                                                                                  \\ \hline
5920                     & 207             & 29                    & 20                      & weka.MultilayerPerceptron                                                                 \\ \hline
5923                     & 265                      & 32                    & 16                      & weka.RandomForest                                                                         \\ \hline
5926                     & 83                       & 32                    & 10                      & weka.SMO\_PolyKernel                                                                      \\ \hline
5978                     & 279             & 31                    & 17                      & weka.AttributeSelectedClassifier\_J48                                                     \\ \hline
6007                     & 248             & 31                    & 15                      & weka.AttributeSelectedClassifier\-\_RandomForest                                            \\ \hline
6073                     & 192             & 32                    & 16                      & weka.AttributeSelectedClassifier\_IBk                                                     \\ \hline
6105                     & 256             & 32                    & 53                      & weka.FilteredClassifier\_RandomForest                                                     \\ \hline
6136                     & 300                      & 30                    & 8                       & weka.FilteredClassifier\_J48                                                              \\ \hline
6140                     & 186                      & 31                    & 19                      & weka.FilteredClassifier\_IBk                                                              \\ \hline
6154                     & 186                      & 31                    & 17                      & weka.FilteredClassifier\-\_AttributeSelectedClassifier\_IBk                                 \\ \hline
6183                     & 64                       & 32                    & 41                      & weka.FilteredClassifier\_SMO\_PolyKernel                                                  \\ \hline
6447                     & 270                      & 30                    & 2                       & weka.FilteredClassifier\-\_AttributeSelectedClassifier\_J48                                 \\ \hline
6458                     & 240                      & 30                    & 3                       & weka.FilteredClassifier\-\_AttributeSelectedClassifier\_RandomForest                        \\ \hline
534                      & 302                      & 65                    & 1                       & weka.Bagging\_SMO\_PolyKernel                                                             \\ \hline
4796                     & 10694                    & 36                    & 3                       & mlr.classif.rpart.preproc                                                                 \\ \hline
5499                     & 20                       & 2                     & 12                      & sklearn.svm.classes.SVC                                                                   \\ \hline
5988                     & 19                       & 8                     & 31                      & weka.AttributeSelectedClassifier\-\_SMO\_PolyKernel                                         \\ \hline
5584                     & 12                       & 2                     & 6                       & sklearn.tree.tree.ExtraTreeClassifier                                                     \\ \hline
189                      & 8                        & 2                     & 1                       & weka.MultiBoostAB\_JRip                                                                   \\ \hline
7064                     & 7                        & 1                     & 8                       & rm.process.polynomial\_by\_binomial\-\_classification\_support\_vector\_machine             \\ \hline
5527                     & 385115                   & 51                    & 8                       & mlr.classif.svm                                                                           \\ \hline
5636                     & 503439                   & 54                    & 6                       & mlr.classif.rpart                                                                         \\ \hline
5859                     & 58809                    & 56                    & 6                       & mlr.classif.rpart                                                                         \\ \hline
5860                     & 3100                     & 27                    & 2                       & mlr.classif.glmnet                                                                        \\ \hline
5890                     & 1740                     & 58                    & 2                       & mlr.classif.kknn                                                                          \\ \hline
5891                     & 44091                    & 51                    & 8                       & mlr.classif.svm                                                                           \\ \hline
5906                     & 2289                     & 24                    & 16                      & mlr.classif.xgboost                                                                       \\ \hline
5965                     & 414678                   & 60                    & 10                      & mlr.classif.ranger                                                                        \\ \hline
5968                     & 918                      & 8                     & 8                       & mlr.classif.ranger                                                                        \\ \hline
5970                     & 68300                    & 55                    & 2                       & mlr.classif.glmnet                                                                        \\ \hline
5971                     & 44401                    & 52                    & 16                      & mlr.classif.xgboost                                                                       \\ \hline
6308                     & 1498                     & 3                     & 2                       & mlr.classif.glmnet                                                                        \\ \hline
6762                     & 4746                     & 7                     & 6                       & mlr.classif.rpart                                                                         \\ \hline
6766                     & 599056                   & 51                    & 2                       & mlr.classif.glmnet                                                                        \\ \hline
6767                     & 491497                   & 52                    & 18                      & mlr.classif.xgboost                                                                       \\ \hline
6794                     & 591831                   & 52                    & 10                      & mlr.classif.ranger                                                                        \\ \hline
6856                     & 2                        & 1                     & 1                       & mlr.classif.randomForest                                                                  \\ \hline
7188                     & 298                      & 35                    & 1                       & mlr.classif.ranger.imputed.dummied.preproc                                                \\ \hline
7189                     & 2324                     & 40                    & 3                       & mlr.classif.rpart.imputed.dummied.preproc                                                 \\ \hline
7190                     & 1215                     & 32                    & 3                       & mlr.classif.RRF.imputed.dummied.preproc                                                   \\ \hline
7607                     & 18686                    & 58                    & 9                       & mlr.classif.ranger                                                                        \\ \hline
7609                     & 41631                    & 59                    & 9                       & mlr.classif.ranger                                                                        \\ \hline
5624                     & 2486                     & 5                     & 7                       & mlr.classif.rpart                                                                         \\ \hline
124                      & 13                       & 2                     & 1                       & weka.MultilayerPerceptron                                                                 \\ \hline
\end{tabular}
\end{table}

\begin{table}[ht]
\begin{tabular}{p{0.10\linewidth} p{0.10\linewidth}  p{0.1\linewidth}  p{0.1\linewidth}  p{0.5\linewidth} }
\hline

153                      & 5                        & 1                     & 1                       & weka.Bagging\_RandomTree                                                                  \\ \hline
243                      & 5                        & 1                     & 1                       & weka.RotationForest\_PrincipalComponents\-\_J48                                             \\ \hline
245                      & 5                        & 1                     & 1                       & weka.RotationForest\_PrincipalComponents\-\_REPTree                                         \\ \hline
246                      & 5                        & 1                     & 1                       & weka.RotationForest\_PrincipalComponents\-\_RandomTree                                      \\ \hline
247                      & 5                        & 1                     & 1                       & weka.RotationForest\_PrincipalComponents\-\_RandomForest                                    \\ \hline
248                      & 5                        & 1                     & 1                       & weka.RotationForest\_PrincipalComponents\-\_LMT                                             \\ \hline
423                      & 10                       & 2                     & 1                       & weka.AdaBoostM1\_SMO\_PolyKernel                                                          \\ \hline
506                      & 34                       & 9                     & 1                       & weka.RotationForest\_PrincipalComponents\-\_J48                                             \\ \hline
633                      & 5                        & 1                     & 1                       & weka.RotationForest\_PrincipalComponents\-\_LMT                                             \\ \hline
2553                     & 4                        & 1                     & 1                       & weka.AttributeSelectedClassifier\-\_CfsSubsetEval\_BestFirst\_IBk                           \\ \hline
4006                     & 8                        & 2                     & 1                       & weka.LibSVM                                                                               \\ \hline
5526                     & 149                      & 7                     & 2                       & mlr.classif.kknn                                                                          \\ \hline
7286                     & 31                       & 1                     & 2                       & mlr.classif.glmnet                                                                        \\ \hline
7290                     & 10                       & 1                     & 4                       & mlr.classif.rpart                                                                         \\ \hline
5626                     & 79                       & 3                     & 17                      & sklearn.ensemble.gradient\_boosting\-.GradientBoostingClassifier                            \\ \hline
5889                     & 1433                     & 20                    & 6                       & mlr.classif.ranger                                                                        \\ \hline
5458                     & 28                       & 4                     & 16                      & mlr.classif.develpartykit.ctree                                                           \\ \hline
5489                     & 30                       & 5                     & 14                      & mlr.classif.develpartykit.ctree                                                           \\ \hline
4828                     & 264                      & 1                     & 10                      & mlr.classif.ksvm                                                                          \\ \hline
214                      & 4                        & 1                     & 1                       & weka.AdaBoostM1\_NaiveBayes                                                               \\ \hline
6741                     & 6                        & 1                     & 21                      & weka.AttributeSelectedClassifier\-\_MultilayerPerceptron                                    \\ \hline
158                      & 5                        & 1                     & 1                       & weka.Bagging\_JRip                                                                        \\ \hline
2566                     & 2                        & 1                     & 2                       & classif.IBk                                                                               \\ \hline
3894                     & 49                       & 4                     & 6                       & mlr.classif.rpart.preproc                                                                 \\ \hline
6003                     & 13537                    & 2                     & 18                      & mlr.classif.xgboost                                                                       \\ \hline
6765                     & 1002                     & 3                     & 2                       & mlr.classif.glmnet                                                                        \\ \hline
5963                     & 809                      & 2                     & 15                      & mlr.classif.xgboost                                                                       \\ \hline
6000                     & 496                      & 1                     & 2                       & mlr.classif.glmnet                                                                        \\ \hline
6322                     & 525                      & 2                     & 8                       & mlr.classif.svm                                                                           \\ \hline
5623                     & 795                      & 3                     & 3                       & mlr.classif.glmnet                                                                        \\ \hline
5969                     & 2543                     & 4                     & 8                       & mlr.classif.svm                                                                           \\ \hline
3737                     & 2                        & 1                     & 2                       & mlr.classif.randomForest                                                                  \\ \hline
829                      & 4                        & 1                     & 4                       & weka.Stacking\_REPTree                                                                    \\ \hline
833                      & 5                        & 1                     & 5                       & weka.Stacking\_ZeroR                                                                      \\ \hline
935                      & 6                        & 1                     & 4                       & weka.LogitBoost\_RandomSubSpace\_REPTree                                                  \\ \hline
6323                     & 30                       & 1                     & 2                       & mlr.classif.kknn                                                                          \\ \hline
7200                     & 811                      & 1                     & 19                      & mlr.classif.xgboost                                                                       \\ \hline
673                      & 96                       & 5                     & 1                       & HIKNN                                                                                     \\ \hline
674                      & 84                       & 5                     & 1                       & ANHBNN                                                                                    \\ \hline
678                      & 96                       & 5                     & 1                       & HwKNN                                                                                     \\ \hline
679                      & 96                       & 5                     & 1                       & KNN                                                                                       \\ \hline
680                      & 83                       & 5                     & 1                       & NHBNN                                                                                     \\ \hline
681                      & 95                       & 5                     & 1                       & HFNN                                                                                      \\ \hline
682                      & 95                       & 5                     & 1                       & DWHFNN                                                                                    \\ \hline
683                      & 95                       & 5                     & 1                       & CBWkNN                                                                                    \\ \hline
684                      & 95                       & 5                     & 1                       & NWKNN                                                                                     \\ \hline
685                      & 95                       & 5                     & 1                       & AKNN                                                                                      \\ \hline

\end{tabular}
\end{table}

\begin{table}[ht]
\begin{tabular}{p{0.10\linewidth} p{0.10\linewidth}  p{0.1\linewidth}  p{0.1\linewidth}  p{0.5\linewidth} }
\hline
688                      & 11                       & 1                     & 1                       & HubMiner.HIKNN                                                                            \\ \hline
689                      & 11                       & 1                     & 1                       & HubMiner.HwKNN                                                                            \\ \hline
690                      & 11                       & 1                     & 1                       & HubMiner.KNN                                                                              \\ \hline
691                      & 4                        & 1                     & 1                       & HubMiner.NHBNN                                                                            \\ \hline
692                      & 11                       & 1                     & 1                       & HubMiner.HFNN                                                                             \\ \hline
693                      & 11                       & 1                     & 1                       & HubMiner.DWHFNN                                                                           \\ \hline
694                      & 11                       & 1                     & 1                       & HubMiner.CBWkNN                                                                           \\ \hline
695                      & 11                       & 1                     & 1                       & HubMiner.NWKNN                                                                            \\ \hline
696                      & 11                       & 1                     & 1                       & HubMiner.AKNN                                                                             \\ \hline
697                      & 11                       & 1                     & 1                       & HubMiner.ANHBNN                                                                           \\ \hline
3994                     & 64                       & 5                     & 1                       & mlr.classif.kknn.preproc.preproc                                                          \\ \hline
5964                     & 506                      & 2                     & 9                       & mlr.classif.ranger                                                                        \\ \hline
5972                     & 120                      & 4                     & 2                       & mlr.classif.kknn                                                                          \\ \hline
6075                     & 495                      & 1                     & 6                       & mlr.classif.rpart                                                                         \\ \hline
2010                     & 7                        & 1                     & 8                       & weka.SimpleLogistic                                                                       \\ \hline
2039                     & 26                       & 2                     & 14                      & weka.AdaBoostM1\_RandomForest                                                             \\ \hline

2073                     & 133                      & 3                     & 13                      & weka.Bagging\_LogitBoost\_DecisionStump                                                   \\ \hline
2277                     & 202                      & 3                     & 15                      & weka.Bagging\_RandomForest                                                                \\ \hline
3489                     & 141                      & 3                     & 12                      & weka.NaiveBayes                                                                           \\ \hline
3490                     & 481                      & 3                     & 29                      & weka.J48                                                                                  \\ \hline
3502                     & 17                       & 2                     & 6                       & weka.ZeroR                                                                                \\ \hline
3960                     & 18                       & 1                     & 15                      & weka.IterativeClassifierOptimizer\_LogitBoost\-\_DecisionStump                              \\ \hline
4289                     & 7                        & 1                     & 20                      & weka.FilteredClassifier\_J48                                                              \\ \hline
5218                     & 6                        & 1                     & 19                      & weka.FilteredClassifier\_RandomForest                                                     \\ \hline
5237                     & 8                        & 1                     & 8                       & weka.FilteredClassifier\_RandomForest                                                     \\ \hline
5253                     & 2                        & 1                     & 2                       & weka.Vote\_RandomForest                                                                   \\ \hline
5295                     & 8                        & 1                     & 13                      & weka.LogitBoost\_DecisionStump                                                            \\ \hline
5301                     & 8                        & 1                     & 15                      & weka.Bagging\_LogitBoost\_DecisionStump                                                   \\ \hline
5315                     & 5                        & 1                     & 9                       & weka.IBk                                                                                  \\ \hline
2614                     & 6                        & 1                     & 2                       & matrixnet\_on\_MagicTelescope\-\_implementation                                             \\ \hline
2629                     & 35                       & 1                     & 6                       & sklearn.ensemble.forest\-.RandomForestClassifier                                            \\ \hline
2793                     & 8                        & 1                     & 12                      & classif.randomForest                                                                      \\ \hline
2799                     & 7                        & 1                     & 7                       & sklearn.ensemble.forest\-.RandomForestClassifier                                            \\ \hline
2823                     & 8                        & 1                     & 11                      & sklearn.ensemble.forest\-.RandomForestClassifier                                            \\ \hline
3414                     & 19                       & 1                     & 2                       & sklearn.ensemble.weight\-\_boosting.AdaBoostClassifier                                      \\ \hline
3425                     & 8                        & 1                     & 9                       & sklearn.neural\_network\-.multilayer\_perceptron.MLPClassifier                              \\ \hline
3434                     & 42                       & 1                     & 19                      & sklearn.ensemble.forest.ExtraTreesClassifier                                              \\ \hline
3439                     & 6                        & 1                     & 3                       & sklearn.ensemble.weight\-\_boosting.AdaBoostClassifier                                      \\ \hline
3442                     & 11                       & 1                     & 5                       & sklearn.neighbors.classification\-.KNeighborsClassifier                                     \\ \hline
5503                     & 14                       & 2                     & 2                       & sklearn.neighbors.classification\-.KNeighborsClassifier                                     \\ \hline
6131                     & 30                       & 1                     & 2                       & mlr.classif.kknn                                                                          \\ \hline
5435                     & 2374                     & 4                     & 3                       & mlr.classif.rpart.preproc                                                                 \\ \hline
7021                     & 11                       & 1                     & 3                       & sklearn.naive\_bayes.BernoulliNB                                                          \\ \hline
5502                     & 6                        & 1                     & 8                       & sklearn.tree.tree.DecisionTreeClassifier                                                  \\ \hline
5521                     & 11                       & 1                     & 3                       & xgboost.sklearn.XGBClassifier                                                             \\ \hline
5604                     & 6                        & 1                     & 12                      & sklearn.ensemble.forest.ExtraTreesClassifier                                              \\ \hline
\end{tabular}
\end{table}

\begin{table}[ht]
\begin{tabular}{p{0.10\linewidth} p{0.10\linewidth}  p{0.1\linewidth}  p{0.1\linewidth}  p{0.5\linewidth} }
\hline

5704                     & 11                       & 1                     & 3                       & DeepForest.DeepForest                                                                     \\ \hline
5788                     & 3                        & 1                     & 3                       & \_\_main\_\_.AUC\_Booster                                                                 \\ \hline
5813                     & 51                       & 1                     & 10                      & \_\_main\_\_.AUCLGBMClassifier                                                            \\ \hline
7680                     & 53                       & 1                     & 2                       & mlr.classif.glmnet                                                                        \\ \hline
7604                     & 20                       & 1                     & 2                       & mlr.classif.kknn                                                                          \\ \hline
5919                     & 6                        & 1                     & 5                       & weka.SimpleLogistic                                                                       \\ \hline
5921                     & 7                        & 1                     & 3                       & weka.Bagging\_J48                                                                         \\ \hline
5922                     & 9                        & 1                     & 10                      & weka.Bagging\_RandomForest                                                                \\ \hline
5960                     & 6                        & 1                     & 7                       & weka.AdaBoostM1\_RandomForest                                                             \\ \hline
6024                     & 43                       & 1                     & 10                      & weka.AttributeSelectedClassifier\_BayesNet                                                \\ \hline
6124                     & 5                        & 1                     & 18                      & weka.FilteredClassifier\_FilteredClassifier\-\_RandomForest                                 \\ \hline
6134                     & 6                        & 1                     & 11                      & weka.J48                                                                                  \\ \hline
6137                     & 7                        & 1                     & 18                      & weka.FilteredClassifier\_NaiveBayes                                                       \\ \hline
6139                     & 4                        & 1                     & 16                      & weka.FilteredClassifier\_RandomTree                                                       \\ \hline
6155                     & 8                        & 1                     & 7                       & weka.FilteredClassifier\_FilteredClassifier\-\_FilteredClassifier\_IBk                      \\ \hline
6156                     & 12                       & 1                     & 1                       & weka.FilteredClassifier\_FilteredClassifier\_IBk                                          \\ \hline
6182                     & 7                        & 1                     & 3                       & weka.Bagging\_FilteredClassifier\_LMT                                                     \\ \hline
6189                     & 54                       & 1                     & 20                      & weka.RandomizableFilteredClassifier\-\_RandomForest                                         \\ \hline
6190                     & 11                       & 1                     & 6                       & weka.RandomizableFilteredClassifier\-\_MultilayerPerceptron                                 \\ \hline
6211                     & 21                       & 1                     & 15                      & weka.FilteredClassifier\_SMO\_RBFKernel                                                   \\ \hline
6212                     & 20                       & 1                     & 4                       & weka.RandomizableFilteredClassifier\-\_SMO\_PolyKernel                                      \\ \hline
6213                     & 7                        & 1                     & 3                       & weka.RandomizableFilteredClassifier\-\_Bagging\_RandomForest                                \\ \hline
6215                     & 13                       & 1                     & 10                      & weka.RandomizableFilteredClassifier\_J48                                                  \\ \hline
6216                     & 8                        & 1                     & 5                       & weka.RandomizableFilteredClassifier\_LMT                                                  \\ \hline
6271                     & 14                       & 1                     & 18                      & weka.FilteredClassifier\_Bagging\-\_AdaBoostM1\_J48                                         \\ \hline
6285                     & 19                       & 1                     & 10                      & weka.FilteredClassifier\_FilteredClassifier\-\_Vote\_RandomForest                           \\ \hline

6309                     & 4                        & 1                     & 7                       & weka.FilteredClassifier\_Bagging\_AdaBoostM1\-\_FilteredClassifier\_J48                     \\ \hline
6345                     & 10                       & 1                     & 20                      & weka.RandomizableFilteredClassifier\-\_RandomTree                                           \\ \hline
6347                     & 13                       & 1                     & 2                       & weka.Bagging\_FilteredClassifier\_J48                                                     \\ \hline
6364                     & 6                        & 1                     & 3                       & weka.RandomizableFilteredClassifier\-\_BayesNet                                             \\ \hline
6365                     & 10                       & 1                     & 4                       & weka.RandomizableFilteredClassifier\-\_NaiveBayes                                           \\ \hline
6376                     & 17                       & 1                     & 7                       & weka.RandomizableFilteredClassifier\-\_SimpleLogistic                                       \\ \hline
6433                     & 16                       & 1                     & 18                      & weka.RandomizableFilteredClassifier\-\_AdaBoostM1\_RandomForest                             \\ \hline
6461                     & 6                        & 1                     & 8                       & weka.NaiveBayesUpdateable                                                                 \\ \hline
6493                     & 14                       & 1                     & 15                      & weka.FilteredClassifier\_FilteredClassifier\-\_DecisionTable                                \\ \hline
6507                     & 20                       & 1                     & 3                       & weka.FilteredClassifier\_FilteredClassifier\-\_RandomizableFilteredClassifier\_RandomForest \\ \hline
\end{tabular}
\end{table}

\begin{table}[ht]
\caption{Number of evaluations per search space in the meta-test split of \benchname{}-v3}
\centering
\begin{tabular}{ccc|ccc}

\hline
\textbf{Search Space ID} & \textbf{Dataset ID} & \textbf{No. Evaluations} & \textbf{Space ID} & \textbf{Dataset ID} & \textbf{No. Evaluations} \\ \hline
4796                     & 3549                & 300                      & 5970                     & 49                  & 3164                     \\ \hline
4796                     & 3918                & 300                      & 5970                     & 34536               & 3286                     \\ \hline
4796                     & 9903                & 300                      & 5970                     & 14951               & 3634                     \\ \hline
4796                     & 23                  & 300                      & 5971                     & 10093               & 2265                     \\ \hline
5527                     & 146064              & 44790                    & 5971                     & 3954                & 2392                     \\ \hline
5527                     & 146065              & 55627                    & 5971                     & 43                  & 3373                     \\ \hline
5527                     & 9914                & 59355                    & 5971                     & 34536               & 3487                     \\ \hline
5527                     & 145804              & 59439                    & 5971                     & 9970                & 3866                     \\ \hline
5527                     & 31                  & 60574                    & 5971                     & 6566                & 4254                     \\ \hline
5527                     & 10101               & 74531                    & 6766                     & 3903                & 39155                    \\ \hline
5636                     & 146064              & 43841                    & 6766                     & 146064              & 43785                    \\ \hline
5636                     & 145804              & 46496                    & 6766                     & 145953              & 47439                    \\ \hline
5636                     & 9914                & 49668                    & 6766                     & 145804              & 52737                    \\ \hline
5636                     & 146065              & 50509                    & 6766                     & 31                  & 60721                    \\ \hline
5636                     & 10101               & 69158                    & 6766                     & 10101               & 66277                    \\ \hline
5636                     & 31                  & 79629                    & 6767                     & 146065              & 44258                    \\ \hline
5859                     & 9983                & 2878                     & 6767                     & 145804              & 45103                    \\ \hline
5859                     & 31                  & 3020                     & 6767                     & 146064              & 46066                    \\ \hline
5859                     & 37                  & 3254                     & 6767                     & 9914                & 47410                    \\ \hline
5859                     & 3902                & 3397                     & 6767                     & 9967                & 48624                    \\ \hline
5859                     & 9977                & 3464                     & 6767                     & 31                  & 68248                    \\ \hline
5859                     & 125923              & 5047                     & 6794                     & 145804              & 52448                    \\ \hline
5860                     & 14965               & 200                      & 6794                     & 3                   & 55460                    \\ \hline
5860                     & 9976                & 268                      & 6794                     & 146065              & 60779                    \\ \hline
5860                     & 3493                & 389                      & 6794                     & 10101               & 65952                    \\ \hline
5891                     & 9889                & 2584                     & 6794                     & 9914                & 69200                    \\ \hline
5891                     & 3899                & 2587                     & 6794                     & 31                  & 102306                   \\ \hline
5891                     & 6566                & 2779                     & 7607                     & 14965               & 647                      \\ \hline
5891                     & 9980                & 2943                     & 7607                     & 145976              & 677                      \\ \hline
5891                     & 3891                & 3083                     & 7607                     & 3896                & 692                      \\ \hline
5891                     & 3492                & 3317                     & 7607                     & 3913                & 696                      \\ \hline
5906                     & 9971                & 254                      & 7607                     & 3903                & 699                      \\ \hline
5906                     & 3918                & 259                      & 7607                     & 9946                & 722                      \\ \hline
5965                     & 145836              & 11228                    & 7607                     & 9967                & 895                      \\ \hline
5965                     & 9914                & 11320                    & 7609                     & 145854              & 1269                     \\ \hline
5965                     & 3903                & 11536                    & 7609                     & 3903                & 1292                     \\ \hline
5965                     & 10101               & 11648                    & 7609                     & 9967                & 1293                     \\ \hline
5965                     & 9889                & 12319                    & 7609                     & 145853              & 1296                     \\ \hline
5965                     & 49                  & 12406                    & 7609                     & 34537               & 1452                     \\ \hline
5965                     & 9946                & 13140                    & 7609                     & 125923              & 1494                     \\ \hline
5970                     & 37                  & 2829                     & 7609                     & 145878              & 1593                     \\ \hline
5970                     & 3492                & 2947                     & 5889                     & 9971                & 298                      \\ \hline
5970                     & 9952                & 3163                     & 5889                     & 3918                & 300                      \\ \hline
\end{tabular}
\end{table}

\end{document}